\documentclass[journal]{IEEEtran}

\usepackage[utf8]{inputenc}
\usepackage[T1]{fontenc}
\usepackage{todonotes}
\usepackage{siunitx}
\usepackage[colorlinks=true, allcolors=blue]{hyperref}
\usepackage{amsmath}

\usepackage{amsfonts}
\usepackage{soul} 
\usepackage{multirow}
\usepackage{makecell}
\usepackage{booktabs} 
\usepackage{adjustbox}
\usepackage{subcaption}
\usepackage[linesnumbered,ruled,vlined]{algorithm2e}
\SetKwInput{kwInit}{Init}
\SetKw{Output}{output}
\SetKwComment{Comment}{$\triangleright$\ }{}

\usepackage{mathtools}
\DeclarePairedDelimiter\ceil{\lceil}{\rceil}

\usepackage{bm}

\usepackage{dblfloatfix}

\usepackage{tikz,xcolor}
\definecolor{lime}{HTML}{A6CE39}
\DeclareRobustCommand{\orcidicon}
{
    \begin{tikzpicture}
    \draw[lime, fill=lime] (0,0) circle [radius=0.16] 
    node[white] {{\fontfamily{qag}\selectfont \tiny ID}};    \draw[white, fill=white] (-0.0625,0.095) circle [radius=0.007];    
    \end{tikzpicture}
    \hspace{0mm}}
\foreach \x in {A, ..., Z}{%
    \expandafter\xdef\csname orcid\x\endcsname{\noexpand\href{https://orcid.org/\csname orcidauthor\x\endcsname}{\noexpand\orcidicon}}
}

\begin{document}


\title{Class Binarization to NeuroEvolution for Multiclass Classification}

\author{Gongjin Lan\hspace{-1mm}\orcidA{}\hspace{-1mm},
Zhenyu Gao$^*$\hspace{-1mm}\orcidC{}\hspace{-1mm}, Lingyao Tong,
Ting Liu$^*$\hspace{-1mm}\orcidD{}
\thanks{This work was partially supported by the Guangdong Natural Science Funds for Young Scholar (No: 2021A1515110641), the National Natural Science Foundation of China (No: 61773197), the Shenzhen Fundamental Research Program (No: JCYJ20200109141622964).
(Corresponding author: Zhenyu Gao and Ting Liu.)}
\thanks{Gongjin Lan is with the Department of Computer Science and Engineering, Southern University of Science and Technology, Shenzhen, 518055, China (e-mail: langj@sustech.edu.cn)}
\thanks{Zhenyu Gao and Ting Liu are with the Department of Computer Science, VU University Amsterdam, the Netherlands (e-mail: vu.gaozhy@gmail.com, t.liu@vu.nl)}
\thanks{Lingyao Tong is with the Department of Clinical, Neuro- \& Developmental Psychology, VU University Amsterdam, the Netherlands (e-mail: l.tong@vu.nl)}
}

\markboth{Journal of \LaTeX\ Class Files,~Vol.
}%
{Shell \MakeLowercase{\textit{et al.}}: Bare Demo of IEEEtran.cls for IEEE Journals}

\maketitle

\begin{abstract}

Multiclass classification is a fundamental and challenging task in machine learning. 
The existing techniques of multiclass classification can be categorized as (i) decomposition into binary (ii) extension from binary and (iii) hierarchical classification.
Decomposing multiclass classification into a set of binary classifications that can be efficiently solved by using binary classifiers, called class binarization, which is a popular technique for multiclass classification.
Neuroevolution, a general and powerful technique for evolving the structure and weights of neural networks, has been successfully applied to binary classification.
In this paper, we apply class binarization techniques to a neuroevolution algorithm, NeuroEvolution of Augmenting Topologies (NEAT), that are used to generate neural networks for multiclass classification.
We propose a new method that applies Error-Correcting Output Codes (ECOC) to design the class binarization strategies on the neuroevolution for multiclass classification.
The ECOC strategies are compared with the class binarization strategies of One-vs-One and One-vs-All on three well-known datasets of \textit{Digit}, \textit{Satellite}, and \textit{Ecoli}.
We analyse their performance from four aspects of multiclass classification degradation, accuracy, evolutionary efficiency, and robustness.
The results show that the NEAT with ECOC performs high accuracy with low variance.
Specifically, it shows significant benefits in a flexible number of binary classifiers and strong robustness.
\end{abstract}


\IEEEpeerreviewmaketitle

\section{Introduction}
\label{sec:introduction}

The classification tasks can be divided into binary (two-class) classification and multiclass classification. 
Multiclass classification is a crucial branch of machine learning, and has been applied in a wide variety of applications, such as medicine, speech recognition, and computer vision.
The existing multiclass classification techniques can be basically divided into three categories, decomposition into binary, extension from binary, and hierarchical classification \cite{aly2005survey}.
Although some classifiers such as Neural Networks (NNs) can classify multiple classes directly as a monolithic multiclass classifier, many state-of-the-art classifiers are inherently proposed for binary classification.
Currently, a popular technique of multiclass classification is to decompose multiclass classification into binary classification \cite{lorena2008review}, which is an efficient method to decode the classification, called class binarization.
The class binarization approaches for multiclass classification have many advantages. 
First, developing binary classifiers is generally much easier than developing multiclass classifiers \cite{galar2011overview}.
Second, many classifiers such as Support Vector Machine (SVM) and C4.5 are inherently proposed for binary classification with outstanding performance \cite{zhou2008data,furnkranz2002round}. 

The binary classifiers (e.g., NNs and SVM) have been successfully applied to the decomposition of multiclass classification.
Neural networks are generally designed by researchers manually.
Using algorithms to automatically generate efficient neural networks is another popular approach for designing neural networks.
Neuroevolution is a popular and powerful technique for evolving the structure and weights of neural networks automatically.
Although neuroevolution approaches have been successfully applied to evolve efficient neural networks for binary classification, it generally struggles to generate neural networks for high accuracy in complex tasks such as multiclass classification \cite{mcdonnell2018divide}. 
In this work, we therefore investigate class binarization techniques in neuroevolution for multiclass classification.

NeuroEvolution of Augmenting Topologies (NEAT) is a popular neuroevolution algorithm that applies evolutionary algorithms (EAs) to generate desired neural networks by evolving both weights and structures \cite{floreano2008neuroevolution}. 
NEAT-based approaches have been successfully applied to a broad range of machine learning tasks such as binary classification \cite{chen2006neuroevolution,lan2019evolving}, regression \cite{hagg2017evolving}, and robotics \cite{lan2021learning}.
However, it is notorious that neural networks evolved by NEAT-based approaches generally suffer severe multiclass classification degradation \cite{mcdonnell2018divide,gao2021neat}.
The performance of neural networks evolved by NEAT degrades rapidly as the number of classes increases \cite{mcdonnell2018divide,lan2019evolving}. 
To solve this issue, we apply the class binarization technique of Error-Correcting Output Codes (ECOC) to decompose multiclass classification into multiple binary classifications that NEAT-based approaches have been successfully applied to.

In general, there are three well-known types of class binarization approaches: One-vs-One (OvO), One-vs-All (OvA), and ECOC \cite{lorena2008review} (see \autoref{sec:binarization}). 
Theoretically, these three approaches work perfectly for multiclass classification when binary classifier predictions are 100\% correct.
However, realistic binary classifiers inevitably make wrong predictions, and these class binarization approaches therefore perform differently for multiclass classification.
Although the class binarization techniques of OvO and OvA have been applied to NEAT-based multiclass classification \cite{mcdonnell2018divide}, it is a novel method that applies ECOC to NEAT for multiclass classification, noted as ECOC-NEAT.

In this work, we mainly concentrate on the two research questions:
1) how ECOC-NEAT performs for multiclass classification? 
2) how the size and quality of ECOC impact the performance of ECOC-NEAT for multiclass classification? 
To answer these two research questions, this study investigates
1) the performance of OvO-NEAT, OvA-NEAT, ECOC-NEAT, and the standard (original) NEAT for multiclass classification,
2) the performance of ECOC-NEAT with different number of classifiers and different ECOCs.
We analyse their performance from four aspects of multiclass degradation, accuracy, training efficiency, and robustness. 

To the convincing conclusions, we choose three popular datasets, (\textit{Digit}, \textit{Satellite}, and \textit{Ecoli}) that are usually used to evaluate the methods in multiclass classification.
The main findings are summarized into two points.
\begin{enumerate}
    \item{ECOC-NEAT offers various benefits compared to the standard NEAT and the NEAT with other class binarization techniques for multiclass classification. 
    \begin{itemize}
        \item{ECOC-NEAT performs comparable high accuracy as OvO-NEAT.}
        \item {ECOC-NEAT outperforms OvO-NEAT and OvA-NEAT in terms of robustness.}
        \item {ECOC-NEAT performs significant benefits in a flexible number of base classifiers.}
    \end{itemize}}
    \item{The size and quality of ECOC greatly influence the performance of ECOC-NEAT.
    \begin{itemize}
        \item{Larger size ECOCs usually contribute to better performance for a given multiclass classification.}
        \item{High quality (optimized) ECOCs perform significantly better than normal ECOCs.}
    \end{itemize}}
\end{enumerate}

The rest of this paper is organized as follows.
In \autoref{sec:related}, we provide an overview of the state-of-the-art studies of class binarization for multiclass classification.
We present the methodology of NEAT and class binarization in \autoref{sec:methodology}. 
Datasets and experimental setup are addressed in \autoref{sec:experiments}. 
We present the results in \autoref{sec:results} from four aspects: multiclass classification degradation, breadth evaluation, evolution efficiency, and robustness.
Finally, we discuss this work in-depth and outlook the future work in \autoref{sec:discussionandfuture}, followed by the conclusions in \autoref{sec:conclusion}.

\section{Related work}
\label{sec:related}

OvO, OvA, and ECOC are three well-known class binarization techniques for multiclass classification. 
Although these three class binarization techniques have been successful applied to many applications, there is a lack of study that applies them (particularly ECOC) to neuroevolution for multiclass classification.

In \cite{ng2014one}, OvA is applied to the diagnosis of concurrent defects with binary classifiers of SVM and C4.5 decision tree.
Adnan and Islam \cite{adnan2015one} applied OvA to the context of Random Forest.
Allwein et al. proposed a general method for combining binary classifiers, in which the ECOC method is applied to a unifying approach with code matrices \cite{allwein2000reducing}.
These studies applied the three class binarization techniques into the traditional classifiers for multiclass classifications.

In the early studies of binary classification in neural networks and neuroevolution, Liu and Yao \cite{liu1999simultaneous} proposed a new cooperative ensemble learning system for designing neural network ensembles, in which a problem is decomposed into smaller and specialized ones, and then each subproblem is solved by an individual neural network.
Abbass et al. \cite{abbass2003pareto} and Garcia-Pedrajas et al. \cite{garcia2005cooperative} presented evolution-based methods to design neural network ensembles.
Lin and Damminda proposed a new algorithm of learning-NEAT that combines class binarization techniques and backpropagation for multiclass classification \cite{chen2006neuroevolution}. 

In the recent study \cite{mcdonnell2018divide}, the class binarization techniques of OvO and OvA are applied to decompose multiclass multiclass into a set of binary classifications for solving the multiclass classification degradation of NEAT, in which binary classifiers are the individual NEAT-evolved neural networks. 
Two ensemble approaches of OvO-NEAT and OvA-NEAT are developed to achieve both higher accuracy and higher efficiency than the standard NEAT. 
Although the class binarization techniques of OvO and OvA have been applied to NEAT for multiclass classification, there is a lack of study that investigates the well-know technique of ECOC in NEAT for multiclass classification.

\section{Methodology}
\label{sec:methodology}

In this section, we describe the neuroevolution algorithm of NEAT, the class binarization techniques of OvO, OvA, and ECOC.

\subsection{NeuroEvolution of Augmenting Topologies}

NEAT is a widely used neuroevolution algorithm that generates neural networks by evolving both weights and structure \cite{floreano2008neuroevolution,stanley2002evolving}.
NEAT evolves neural networks with flexible topology, starting from the elementary topology where all input nodes are connected to all output nodes, and adding nodes and connections via the operations of recombination and mutations, which leads to an augmented topology.
In this work, NEAT is also allowed to delete nodes as well as connections.
NEAT searches optimal neural networks through weight space and topological space simultaneously. 
There is no need for an initial or pre-defined fixed-topology that relies on the experience of researchers.
Recombination and mutation induce an optimal topology of NN to an effective network. 

\begin{figure}[ht!]
\centering 
\includegraphics[width=0.3\textwidth]{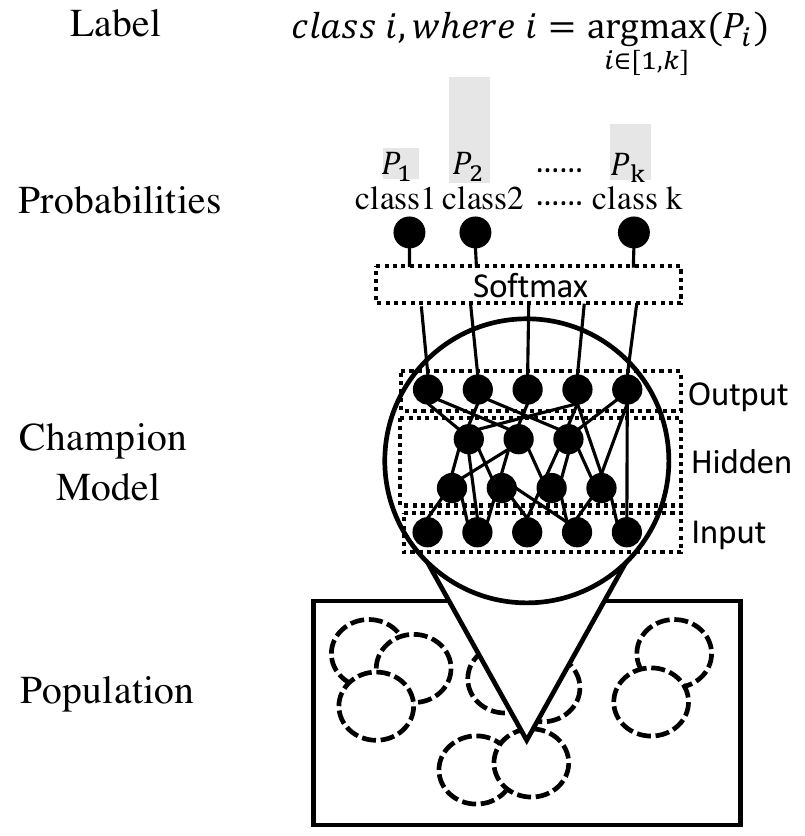} 
\caption{Illustration of evolving neural networks by NEAT for multiclass classification} 
\label{fig:standardneat} 
\end{figure}

An example of evolving neural networks by NEAT for multiclass classification is illustrated in \autoref{fig:standardneat}. 
NEAT aims to generate an optimal neural network (i.e., highest fitness) as the winning multiclass classifier.
In particular, NEAT generates a binary classifier when the number of classes is two, where the NEAT is referred to as binary-NEAT (B-NEAT) as shown in the left part of \autoref{fig:ecocneat}.
The number of nodes of the input layer is the dimensions of feature ($\mathcal{D}$), and the number of output nodes is the number of classes ($k$). 
We apply a softmax operation in the final layer to output probabilities of each class for multiclass classification. 
The class with the highest probability is predicted as the result.

NEAT is essentially a variant of evolutionary algorithms.
Therefore, the fitness function is crucial to guide the convergence of evolving desired neural networks.
In this work, we evaluate the performance of evolved neural networks with the prediction accuracy, that is the percentage of correct predictions.
We note the number of correct predictions as $\mathcal{N}_c$, the number of total predictions as $\mathcal{N}_t$.
The fitness ($f$) can be calculated as $f  = \mathcal{N}_c/\mathcal{N}_t$.

Although NEAT can directly evolve neural networks for multiclass classification, it suffers the notorious multiclass classification degradation \cite{chen2006neuroevolution}. 
We apply NEAT as the baseline method for multiclass classification in this study, i.e., standard NEAT.

\subsection{Class Binarization}
\label{sec:binarization}

\subsubsection{One-vs-One}

The class binarization of OvO (also called All-vs-All) technique converts $k$-class classification into $\binom{k}{2}$ binary classifications that are constructed by using the class $i ~(i = 1, ..., k-1)$ as the positive examples and other classes $j > i ~(j=2, ..., k)$ as the negative examples \cite{aly2005survey}.
That is, each class is compared with each other class separately.
The existing studies \cite{allwein2000reducing,hsu2002comparison} show that OvO generally performs better than OvA approaches.

\begin{figure*}[ht!] 
\centering
\includegraphics[width=0.9\textwidth]{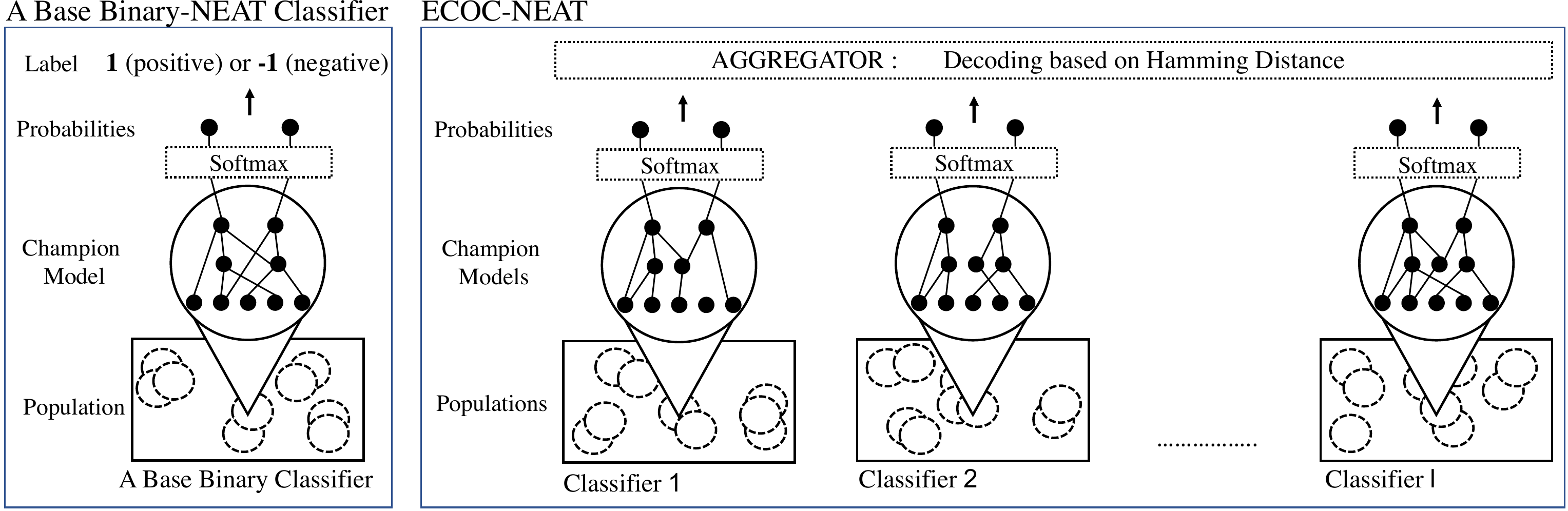} 
\caption{ECOC-NEAT for multiclass classification. The left part shows an evolved base (binary) classifier by NEAT. The right part shows the ECOC-NEAT with base classifiers.}
\label{fig:ecocneat} 
\end{figure*}

NEAT evolves neural networks as binary classifiers for each binary classification.
An example of evolving binary classifiers (base classifiers) by NEAT is shown in the left part of \autoref{fig:ecocneat}.
The voting strategy is usually used to fuse these binary classifications for multiclass classification. 
Each binary classifier votes to one class, and the class with the highest votes is predicted as the result. 
The OvO technique and base classifiers evolved by NEAT are combined for multiclass classification, i.e., OvO-NEAT. 
Although NEAT are effective to generate binary classifiers, the OvO-NEAT technique requires building a large number of $\binom{k}{2}$ base classifiers. 

\subsubsection{One-vs-All} 
OvA (also called One-vs-Rest or One-against-All) technique converts a $k$-class classification into $k$ binary classifications. 
These binary classifications are constructed by using class $i$ as the positive examples and the rest of classes $j~ (j=1, ..., k, j \neq i)$ as the negative examples.
Each binary classifier is used to distinguish class $i$ from all the other $k-1$ classes.
When testing an unknown example, the class with maximum prediction is considered the winner \cite{aly2005survey}.
Compared to OvO, OvA provides considerable performance but requires fewer ($k$) classifiers. 

\subsubsection{Error-Correcting Output Codes}
\label{sec:test}

ECOC is a class binarization method for multiclass classification, inspired by error-correcting code transmission techniques from communications theory \cite{dietterich1994solving}. 
It encodes $\mathcal{N}$ binary classifiers to predict $k$ classes. 
Each class is given an $\mathcal{N}$-length codeword according to an ECOC matrix $\mathbb{M}$. 
Each codeword in $\mathbb{M}$ is mapped to a certain class.
An example of ECOC for $k = 4$ classes and $\mathcal{N} = 7$-bit codewords are shown in \autoref{tab:ecoc}. 
Each column is used to train a binary classifier. 
When testing an unseen class, the codeword predicted by $\mathcal{N}$ classifiers is matched to the $k$ codewords in $\mathbb{M}$.
In this work, we adopt hamming distance to match predicted codeword and the ECOC codewords.
The class with the minimum hamming distance is considered as the predicted class.

\begin{table}[!ht]
    \centering \small
    \setlength\tabcolsep{6pt} 
    \caption{An example of ECOC for $k = 4$ classes with a size of $\mathcal{N} = 7$ bit codewords.}
    \begin{tabular}{c c c c c c c c } \toprule
        \multirow{2}{*}{classes} & \multicolumn{7}{c}{classifers} \\ \cline{2-8}
           & $f_1$ & $f_2$ & $f_3$ & $f_4$ & $f_5$ & $f_6$ & $f_7$ \\ \midrule
        $c_1$ & 1 & 1 & 1 & 1 & 1 & 1 & 1 \\
        $c_2$ & 0 & 0 & 0 & 0 & 1 & 1 & 1 \\
        $c_3$ & 0 & 0 & 1 & 1 & 0 & 0 & 0 \\
        $c_4$ & 0 & 1 & 0 & 1 & 0 & 1 & 0 \\ \bottomrule
    \end{tabular}
    \label{tab:ecoc}
\end{table}

Unlike OvO and OvA methods that convert a multiclass classification into a fixed number of binary classifications, ECOC allows each class to be encoded with a flexible number of binary classifications, and allows extra models to act as overdetermined predictions that can result in better predictive performance \cite{rokach2010pattern}.
The row of ECOC needs to be a unique codeword, and columns are neither identical nor complementary.
In ECOC, the size of codewords (rows) is the number of classes, and thus the size of ECOC refers to the number of base classifiers in this work.
The larger size ECOC provides more bits to correct errors, but too many classifiers cause redundancy which costs a lot of computation in training and classification. 

For $k$ classes, the minimum size of ECOC is $\ceil{log_2 k}$. 
For example, 10 classes require a minimum size of 4 bits ECOC that are sufficient for representing each class with a unique codeword.
We call the ECOC with a minimum size of $\mathcal{N} = \ceil{log_2 k}$ as minimal ECOC.
The maximum size of ECOC is $2^{k-1} -1$ for $k$ classes. 
The ECOC with a maximum size is generally called as exhaustive ECOC \cite{dietterich1994solving}. 
The upper and lower bounds of ECOC size can be expressed as:
\begin{equation}
\ceil{log_2 k} \leq \mathcal{N} \leq 2^{k-1} -1, ~\mathcal{N} \in \mathbb{Z} 
\label{equ:sizeecoc}
\end{equation}
where $\mathbb{Z}$ is the positive integer set.
Besides OvO, OvA, minimal ECOC, and exhaustive ECOC, the mid-length ECOC is another representative class binarization technique with intermediate-length code whose size is $\mathcal{N} = \ceil{10\log_2(k)}$ \cite{allwein2000reducing}.

\begin{figure}[!ht] 
\centering 
\includegraphics[width=0.4\textwidth]{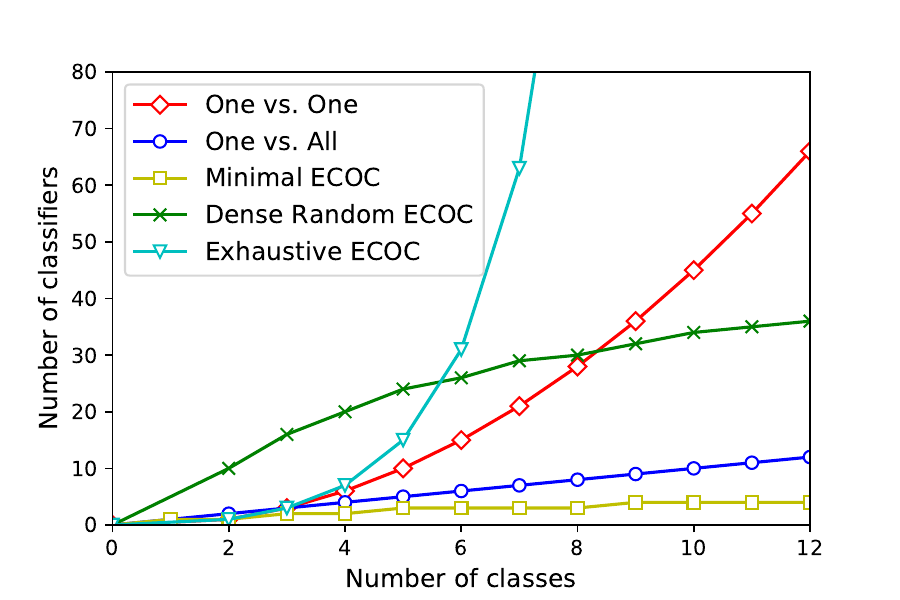}
\caption{Number of classifiers over the number of classes for class binarization techniques.} 
\label{fig:numberofclassifiers} 
\end{figure}

The number of base classifiers varies as the number of classes increases using these class binarization is shown in \autoref{fig:numberofclassifiers}. 
OvO requires a polynomial number of base classifiers ($O(k^2)$). 
However, OvA needs a linear number of classifiers ($O(k)$). 
For minimal ECOC and mid-length ECOC, $O(\log(k))$ binary classifiers are required. 
The number of base classifiers used in exhaustive ECOC is exponential ($O(2^k)$).

\SetKw{Output}{Output}
\newcommand{\argmax}{\arg\!\max}
\newcommand{\argmin}{\arg\!\min}

\begin{algorithm}[!ht] \small
\KwData{Construct an ECOC $\mathbb{M}$ with $\mathcal{N}$ columns and the corresponding positive dataset $\mathcal{S}_{j}$ and negative dataset $\bar{\mathcal{S}}_{j}$ for each base classifier, where $j \in [1,\mathcal{N}]$; Test dataset $\mathcal{X}$;
Initialize binary classifier set $\mathbb{F}(f_1, f_2, ..., f_{\mathcal{N}})$.}
\Output{$\mathcal{N}$ binary classifiers; 
predictions $\mathcal{Y}$ for each test sample $x$ in $\mathcal{X}$}. \\
\tcp{Generating base classifiers by NEAT}
\ForEach {$j \in [1,\mathcal{N}]$}
{
\While{$i <= \mathcal{G}/\mathcal{N}$, $\mathcal{G}$ is total generations.}
{
    binary-NEAT evolves neural networks $\mathbb{N}_i$ for predicting the data from $\mathcal{S}_{j}$ and $\bar{\mathcal{S}}_{j}$. 
}
$f_j = \argmax (\mathbb{N}_i)$. \\
Update $\mathbb{F}(f_1, f_2, ..., f_{\mathcal{N}})$.
}
\tcp{multiclass classification}
\ForEach{$x \in \mathcal{X}$}  
{
\tcp{binary classification of $\mathcal{N}$ base classifiers on each test sample $x$.}
$\mathbb{F}(x)\leftarrow\{f_1({x}),f_2({x}),\cdots,f_{\mathcal{N}}({x})\}$ 
\tcp*[l]{multiclass classification by hamming distance.}
$\mathcal{Y} \leftarrow \argmin_r \Delta(\mathbb{M}_r,\mathbb{F}(x)), r \in [1,k]$;
}
\caption{ECOC-NEAT for multiclass classification}
\label{alg:ecoc_neat}
\end{algorithm}

The exhaustive ECOC is not generally applied to the multiclass classifications with a large number of classes because it requires too many binary classifiers.
The mid-length ECOC can be constructed by choosing codewords from exhaustive ECOC to satisfy row and column separation conditions.
$\mathcal{N}$ columns are randomly chosen from an exhaustive code to construct the random code matrix when the number of binary classifiers is $\mathcal{N}$.
For example, if $k=4, \mathcal{N} = 3$, we can choose $f_1$, $f_2$, and $f_3$ from the exhaustive ECOC (\autoref{tab:ecoc}) to construct a mid-length ECOC.
By contrast, we cannot choose $f_5$, $f_6$, and $f_7$ because in that case the codeword of $c_1$ will be exactly the same as the codeword of $c_2$, in which the class $c_1$ and $c_2$ can not be classified.

In general, optimized ECOC performs better than normal ECOC \cite{bautista2012minimal} at the same size. 
In this work, we investigate whether optimized minimal-ECOC outperforms minimal ECOC (see \autoref{sec:comparison}). 
NEAT evolves neural networks to constitute a set of binary classifiers.
Hamming distance is used to determine the final prediction.
The pseudo-code of ECOC-NEAT is shown in \autoref{alg:ecoc_neat}.

\section{Experiments}
\label{sec:experiments}

In this section, we introduce the datasets, hyperparameter configurations, implementation, and the measurements.

\subsection{Datasets}
\label{sec:datasets}

In this work, we choose the three well-known datasets of \textit{Digit} from the ski-learn package \cite{scikit-learn}, \textit{Satellite} and \textit{Ecoli} from the machine learning repository of the University of California, Irvine (UCI) \cite{asuncion2007uci}.
These three datasets with high quality data are prevalent and widely used in multiclass classification tasks.
The properties of these three datasets are summarized in \autoref{tab:dataset}.

\begin{table}[!ht]
\centering \small
\renewcommand{\arraystretch}{1.0}
\setlength\tabcolsep{3pt} 
\caption{The properties of three popular datasets of \textit{Digit}, \textit{Satellite} and \textit{Ecoli}.} 
\begin{tabular}{l c c c c}
\toprule
Dataset   & \makecell[l]{Training \\ samples} & Test samples & classes $(k)$ & \makecell[l]{Dimensions \\ of feature} \\ \midrule
Digit     & $1,617$ & $180$  & 10  & 64 \\ 
Satellite & 4,435   & 2,000  & 6  & 36  \\ 
Ecoli   & 336     & 10-fold   &  8  & 7 \\ \bottomrule
\end{tabular}
\label{tab:dataset} 
\end{table}

\subsection{Experimental Setup}
\label{subsec:experiments}

This work compares the newly proposed ECOC-NEAT with the standard NEAT, OvO-NEAT, OvA-NEAT, and ECOC-NEAT. 
A hyper-parameter configuration of NEAT is summarized in \autoref{tab:parameters} which are the same for evolving binary classifiers on the three datasets.
The dimensions of the input layer for evolved binary classifiers equal the dimensions of feature for a dataset (the last column in \autoref{tab:dataset}).
The dimension of outputs in NEAT is set to $2$ for evolving binary classifiers.
In the standard NEAT, the dimension of outputs equals the number of classes $k$ for multiclass classification.

\begin{table}[!ht]
\centering \small
\renewcommand{\arraystretch}{1.0}
\setlength\tabcolsep{2pt} 
\caption{The parameter configurations of NEAT.}
\begin{tabular}{l c | l c}
\toprule
parameters & value & parameters & value \\ \midrule
pop\_size & 200 & weight\_mutate\_rate & 0.8 \\ 
elitism & 2 & activation\_mutate\_rate & 0.3 \\
initial\_connection & 0.1 & conn\_delete\_prob & 0.1 \\ 
conn\_add\_prob & 0.8 & node\_delete\_prob & 0.1 \\
node\_add\_prob & 0.7 & bias\_mutate\_rate & 0.7 \\
survival\_threshold & 0.2 & max\_fitness\_threshold & 1.0 \\
max\_stagnation & 15 & compatibility\_threshold & 2.5 \\ 
elite\_species & 3 & compatibility\_weight\_coefficient & 0.6 \\
feed\_forward & true & compatibility\_disjoint\_coefficient & 1.0 \\ \bottomrule
\end{tabular}
\label{tab:parameters} 
\end{table}

We set the number of generations as $\mathcal{G} = 3000$ for each evolution process of the standard NEAT.
For a fair comparison, we apply the same total number of generations ($\mathcal{G} = 3000$) to evolve binary classifiers for these class binarization techniques.
Specifically, each base classifier is generated by an evolution of ($\mathcal{G}/\mathcal{N}$) generations in NEAT if there are $\mathcal{N}$ classifiers for a class binarization technique.

We implement the standard and binary NEAT based on an open-source NEAT-Python \footnote{\url{https://github.com/CodeReclaimers/neat-python}}. 
The experiments are run on the computer with a dual 8-core 2.4 GHz CPU (Intel Haswell E5-2630-v3) and 64 GB memory. 

\section{Results}
\label{sec:results}

We show the results from the following four aspects: multiclass classification degradation, breadth evaluation, evolution efficiency, and robustness.

\subsection{Multiclass Classification Degradation}
\label{sec:multiclass}

The accuracy of multiclass classification generally decreases as the number of classes increases due to the task that becomes more difficult.
We test the multiclass classification degradation of NEAT (including the standard NEAT and NEAT with class binarization) on the \textit{Digit} dataset, in which the number of classes varies from two to ten. 
For example, the two-class and three-class classification predicts the digit "0, 1" and "0, 1, 2" respectively.

\subsubsection{Multiclass Classification Degradation of the Standard NEAT}
\begin{figure}[!ht]
\centering
\includegraphics[width=0.4\textwidth]{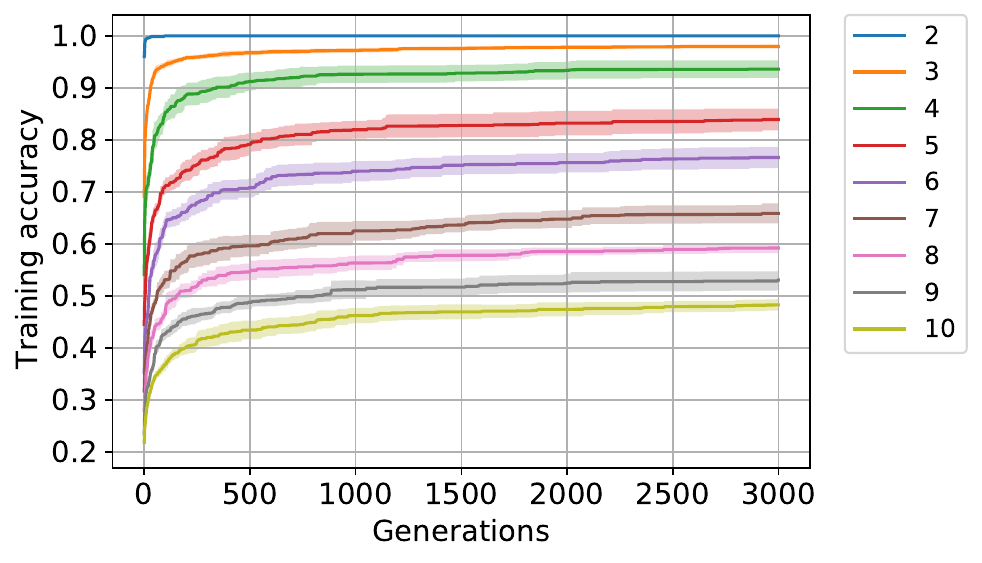}
\caption{The convergence processes of NEAT for the multiclass classification from two to ten classes. The shadows show 95\% confidence intervals.}
\label{fig:multiclass}
\end{figure}

The standard NEAT is used to evolve neural networks for the classification from two classes to ten classes.
The experiments are repeated ten times on the \textit{Digit} dataset.
The convergence processes of the standard NEAT are shown in \autoref{fig:multiclass} where we presents the training accuracy over generations during the evolution of neural networks with 2-10 classes.

The results clearly show that the accuracy decreases dramatically as the number of classes increases.
The classification of two and three classes quickly converges to the high accuracy of more than 95\% with narrow confidence intervals which means their evolution processes are steady.
However, the accuracy converges to the catastrophic value for the classifications with many classes.
In particular, the 10-classes classification (yellow line) converges to an accuracy of less than 50\% slowly. 
In summary, the results show that NEAT performs well for the classification with a few classes (particularly binary classification), but its performance significantly degrades over the number of classes increases.

\subsubsection{Multiclass Classification Degradation of NEAT with Class Binarization}

\begin{figure}[!ht] 
\centering 
\includegraphics[width=0.42\textwidth]{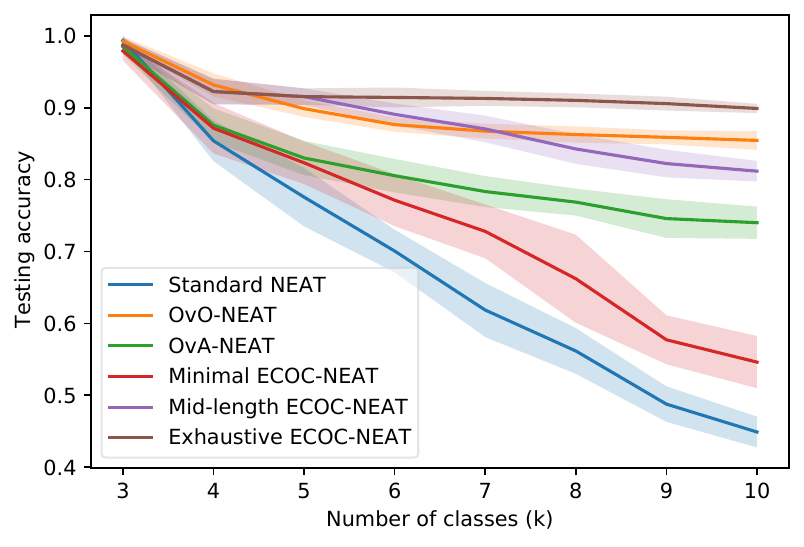} 
\caption{Testing accuracy over number of classes for the multiclass classification methods.} 
\label{fig:degradation} 
\end{figure}

We investigate the degradation of the standard NEAT, OvO-NEAT, OvA-NEAT, and three different sizes of ECOC-NEAT (including minimal ECOC-NEAT, mid-length NEAT, exhaustive ECOC-NEAT) for multiclass classification. 
\autoref{fig:degradation} presents the performance of the standard NEAT, OvO-NEAT, OvO-NEAT, and three ECOC-NEAT for multiclass classifications with a varying number of classes from three to ten.

\begin{table*}[!ht]
\centering \small
\renewcommand{\arraystretch}{1.0}
\setlength\tabcolsep{4pt} 
\caption{Comparison of different methods on the three datasets of \textit{Digit} (10 classes), \textit{Satellite} (6 classes), and \textit{Ecoli.} (8 classes). Each method is run ten times and token an average of results. 
The total generation of each method is identical $\mathcal{G} = 3000$. 
$\mathcal{N}$-bit ECOC-NEAT represents different sizes of mid-length ECOC-NEAT that is $\mathcal{N}$ base classifiers. $\overline{\mathcal{A}_b}$ represents the average training accuracy of binary classifiers.}
\begin{tabular}{l l l c c c c c c}
\toprule
Dataset & \multicolumn{2}{c}{Method} & \makecell[l]{Number of \\ classifiers} & \makecell[l]{Testing \\ accuracy} & Variance & \makecell[l]{Training \\ accuracy} & $\overline{\mathcal{A}_b}$ & \makecell[l]{Average training \\ time/Generation(s)} \\ \midrule
\multirow{9}{*}{Digit} & \multirow{3}{*}{Existing} & Standard NEAT & 1 & 0.449 & 9.56$\times10^{-4}$ & 0.484 & 0.484 & 13.74 \\
& & OvO-NEAT & 45 & \underline{0.866} & 4.94$\times10^{-4}$  & \textbf{\colorbox{lightgray!60}{0.953}} & \textbf{\colorbox{lightgray!60}{0.989}} &  \textbf{\colorbox{lightgray!60}{0.99}} \\ 
& & OvA-NEAT & 10 & \underline{0.740} & 10.46$\times10^{-4}$ & 0.820 & 0.976 & 8.40 \\ \cline{2-9}
& \multirow{6}{*}{Ours} & Minimal ECOC-NEAT & 4 & \underline{0.535} & 72.01$\times10^{-4}$ & 0.614 & 0.865 & 10.49 \\ \cline{3-9}
& & 10-bit ECOC-NEAT & 10 & \underline{0.651} & 15.78$\times10^{-4}$ & 0.724 & 0.860 & 8.49 \\
& & 45-bit ECOC-NEAT & 45 & \underline{0.819}  & 9.04$\times10^{-4}$ & 0.876 & 0.837 & 5.62 \\
& & 100-bit ECOC-NEAT & 100 & \underline{0.845} & 6.77$\times10^{-4}$ & 0.894 & 0.812 & 4.96 \\
& & 250-bit ECOC-NEAT & 250 & \underline{0.876}  & 2.76$\times10^{-4}$ & 0.908 & 0.793 & 4.60 \\ \cline{3-9}
& & Exhaustive ECOC-NEAT & 511 & \underline{\underline{{\textbf{\colorbox{lightgray!60}{0.899}}}}} & \textbf{\colorbox{lightgray!60}{0.95$\times10^{-4}$}} & 0.909 & 0.783 & 4.53 \\ \bottomrule
\multirow{9}{*}{Satellite} & \multirow{3}{*}{Existing} & Standard NEAT & 1 &  0.754 & 0.99$\times10^{-4}$ & 0.774 & 0.774 & 5.09\\
& & OvO-NEAT & 28 & \underline{0.842} & 0.79$\times10^{-4}$ &  \textbf{\colorbox{lightgray!60}{0.914}} & \textbf{\colorbox{lightgray!60}{0.989}} &   \textbf{\colorbox{lightgray!60}{0.12}} \\ 
& & OvA-NEAT & 8 & \underline{0.787} & 1.53$\times10^{-4}$ & 0.848 & 0.979 & 0.75\\ \cline{2-9}
& \multirow{6}{*}{Ours} & Minimal ECOC-NEAT & 3 & \underline{0.765} & 2.28$\times10^{-4}$ & 0.816  & 0.922 & 1.11 \\ \cline{3-9}
& & 8-bit ECOC-NEAT & 8 & \underline{0.790} & 3.24$\times10^{-4}$ & 0.844  & 0.926 & 0.94 \\
& & 15-bit ECOC-NEAT & 15 & \underline{0.828} & 3.90$\times10^{-4}$ & 0.870 & 0.922 & 0.84 \\
& & 28-bit ECOC-NEAT & 28 & \underline{\underline{\textbf{\colorbox{lightgray!60}{0.849}}}} & 0.79$\times10^{-4}$ & 0.881 & 0.917 & 0.73 \\
& & 40-bit ECOC-NEAT & 40 & \underline{\underline{0.848}} & \textbf{\colorbox{lightgray!60}{0.46$\times10^{-4}$}} & 0.885 & 0.914 & 0.68 \\
& & 60-bit ECOC-NEAT & 60 & \underline{0.848} & 2.16$\times10^{-4}$ & 0.885 & 0.910 & 0.62 \\ \cline{3-9}
& & Exhaustive ECOC-NEAT & 127 & \underline{0.837} & 0.88$\times10^{-4}$ & 0.873  & 0.900 & 0.55 \\ \bottomrule
\multirow{9}{*}{Ecoli.} & \multirow{3}{*}{Existing} & Standard NEAT & 1 &  0.754 & 0.99$\times10^{-4}$ & 0.774 & 0.774 & 5.09\\
& & OvO-NEAT & 28 & \underline{0.842}  & 0.79$\times10^{-4}$ &  \textbf{\colorbox{lightgray!60}{0.914}} & \textbf{\colorbox{lightgray!60}{0.989}}  &   \textbf{\colorbox{lightgray!60}{0.12}} \\ 
& & OvA-NEAT & 8 & \underline{0.787}  & 1.53$\times10^{-4}$ & 0.848  & 0.979  & 0.75\\ \cline{2-9}
& \multirow{6}{*}{Ours} & Minimal ECOC-NEAT & 3 & \underline{0.765}  & 2.28$\times10^{-4}$ & 0.816  & 0.922 & 1.11 \\ \cline{3-9}
& & 8-bit ECOC-NEAT & 8 & \underline{0.790} & 3.24$\times10^{-4}$ & 0.844 & 0.926 & 0.94 \\
& & 15-bit ECOC-NEAT & 15 & \underline{0.828} & 3.90$\times10^{-4}$ & 0.870 & 0.922 & 0.84 \\
& & 28-bit ECOC-NEAT & 28 & \underline{\underline{\textbf{\colorbox{lightgray!60}{0.849}}}} & 0.79$\times10^{-4}$ & 0.881 & 0.917 & 0.73 \\
& & 40-bit ECOC-NEAT & 40 & \underline{\underline{0.848}} & \textbf{\colorbox{lightgray!60}{0.46$\times10^{-4}$}} & 0.885 & 0.914 & 0.68 \\
& & 60-bit ECOC-NEAT & 60 & \underline{0.848} & 2.16$\times10^{-4}$ & 0.885 & 0.910 & 0.62 \\ \cline{3-9}
& & Exhaustive ECOC-NEAT & 127 & \underline{0.837} & 0.88$\times10^{-4}$ & 0.873 & 0.900 & 0.55 \\ \bottomrule
\end{tabular}
\label{tab:result3dataset} 
\end{table*}
The results show that not only the resulting accuracy of the standard NEAT decreases dramatically but also that of NEAT with class binarization techniques decreases as the number of classifications increases.
Importantly, the methods of NEAT with class binarization techniques perform slighter decreases than the standard NEAT.
In particular, exhaustive ECOC-NEAT, OVO-NEAT, and mid-length ECOC-NEAT perform remarkable robustness over the number of classes increases.
Moreover, they exhibit higher accuracy and less variance than the standard NEAT.
The mid-length ECOC-NEAT with a moderate number of base classifiers provides competitive performance compared to OvO-NEAT and the exhaustive ECOC-NEAT that requires a large number of base classifiers. 
The exhaustive ECOC-NEAT outperforms the mid-length ECOC-NEAT that outperforms minimal ECOC-NEAT.
We summarize that ECOC-NEAT methods with large size ECOC (i.e., a large number of base classifiers) generally tends to perform better than small size ECOC.
Intriguingly, minimal ECOC-NEAT with a few bases learners still significantly performs better than the standard NEAT for multiclass classification. 

\subsection{Comprehensive Comparison}
\label{sec:comparison}

We investigate the standard NEAT, OvO-NEAT and OvA-NEAT and the proposed ECOC-NEAT methods with different codes including the minimal, mid-length and exhaustive code on the three datasets. 
Specially, we apply the mid-lengths ECOC-NEAT with different sizes to investigate the relationship between the size of ECOC-NEAT and their resulting accuracy. 
The performance of these methods is shown in \autoref{tab:result3dataset} where we presents 1) testing accuracy (accuracy on test set), 2) variance of testing accuracy over ten repetitions, 3) training accuracy on the training set, 4) average training accuracy of each base classifier, and 5) average training time per generation.

The results show that NEAT with class binarization techniques significantly outperform the standard NEAT in terms of accuracy.
ECOC-NEAT even the minimal ECOC-NEAT performs higher accuracy than the standard NEAT on the three datasets.
The exhaustive ECOC-NEAT with the largest number of base classifiers performs the smallest variances that represent the strong robustness.
Conversely, the minimal ECOC-NEAT with a few binary classifiers performs large variances that mean the fluctuating performance. 

The average training accuracy of each base classifier shows the performance of each evolved binary classifier on the training dataset.
The binary classifiers in OvO-NEAT perform the best average training accuracy because it decomposes multiclass classifications into simple binary classification tasks.
The evolved binary classifiers in ECOC-NEAT methods perform lower average accuracy than OvO-NEAT and OvA-NEAT because the binary classifications in ECOC-NEAT are generally challenging and each classifier in ECOC-NEAT is assigned a few $(\mathcal{G}/\mathcal{N})$ generations to evolve.
However, the ECOC-NEAT methods still perform high accuracy for multiclass classification due to the high quality of ensemble in ECOC. 

NEAT in these methods takes different computation times to evolve binary classifiers. 
The standard NEAT takes much more computation time per generation to evolve classifiers than the NEAT with class binarization techniques. 

\subsection{Size of ECOC-NEAT}
\label{ssec:lengthofecoc}
\begin{figure}[!ht]
\centering
\centering
\includegraphics[width=0.4\textwidth]{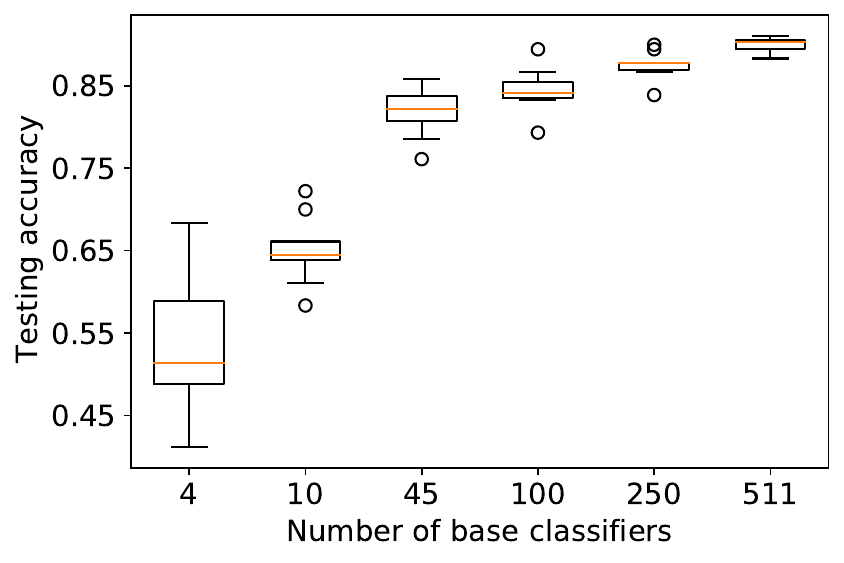}
\caption{Testing accuracy over ECOC-NEAT size on \textit{Digit}.} 
\label{fig:numberofbaselearners2}
\end{figure}

The size of ECOC performs a significant influence on their performance for multiclass classification  \cite{berger1999error}.
To further observe the influence of the size of ECOC on their performance, we visualize the testing accuracy and variance over the size of ECOC (the results in \autoref{tab:result3dataset}) in \autoref{fig:numberofbaselearners2}.
The visualization shows that testing accuracy increases as the number of base classifiers increases and the small size ECOC-NEAT performs fluctuating testing accuracy.
A similar observation can be illustrated from the results on the \textit{Satellite} and \textit{Ecoli} datasets (\autoref{tab:result3dataset}). 

\subsection{Quality of ECOC-NEAT}
\label{subsec:codedesign}

Besides the size of ECOC, the quality of ECOC is another crucial factor for the performance of ECOC-NEAT.
The minimal ECOC-NEAT with a few base classifiers generally perform sensitive to the quality of ECOC.
Thus, we concentrate on the quality of the minimal ECOC-NEAT.

\subsubsection{on the \textit{Satellite} Dataset}

\begin{figure}[!ht]
\centering 
\includegraphics[width=0.45\textwidth]{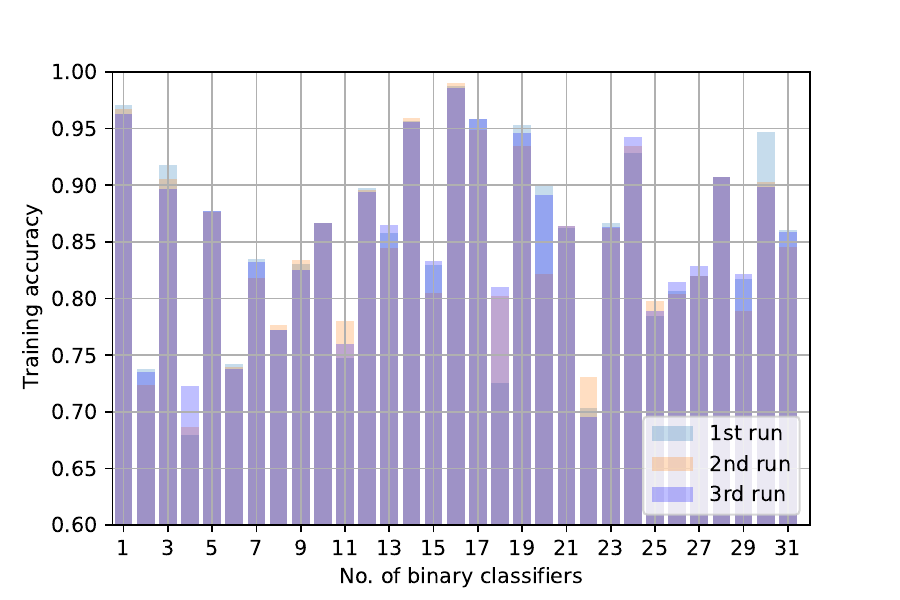} 
\caption{Training accuracy of the 31 binary classifiers in an exhaustive ECOC-NEAT on the \textit{Satellite} dataset for three repetitions.} 
\label{fig:accueach_each_satellite}
\end{figure}

ECOC-NEAT with high training accuracy binary classifiers generally performs high testing accuracy for multiclass classification.
The binary classification tasks in an ECOC-NEAT are generally with various difficulty.
The exhaustive ECOC for the \textit{Satellite} dataset with $k=6$ classes is with 31 columns (see \autoref{tab:result3dataset}).
We run an exhaustive ECOC-NEAT to evolve the 31 binary classifiers on the \textit{Satellite} dataset for three repetitions.
The training accuracy of these 31 binary classifiers is shown in the bar chart of \autoref{fig:accueach_each_satellite}.
The results show that these binary classifiers in the exhaustive ECOC-NEAT perform significant different accuracy from around 70\% to 98\%.

\begin{figure*}[!ht]
\centering \hspace{-0.1cm}
    \begin{subfigure}[t]{.33\textwidth}
        \centering
        \includegraphics[width=\textwidth]{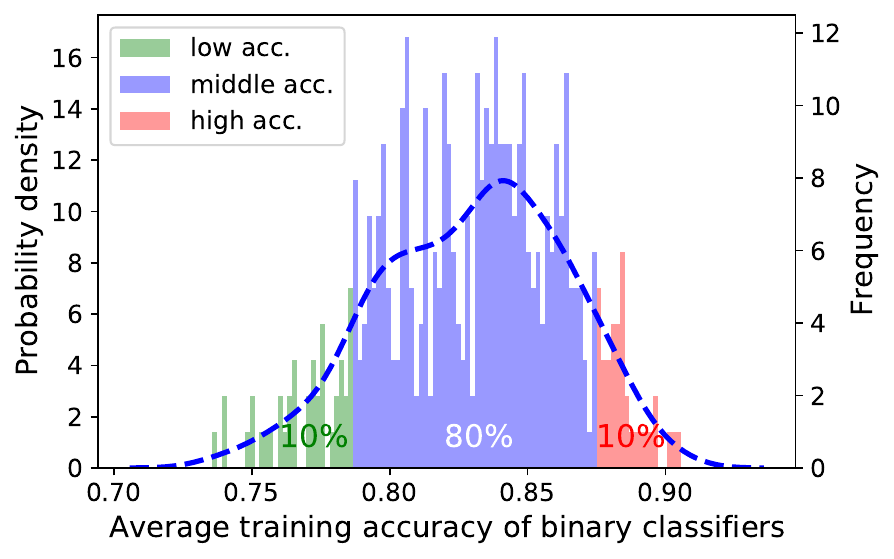}
        \caption{on the \textit{Satellite} with 6 classes}
        \label{fig:distribution_satellite}
    \end{subfigure} \hspace{-0.2cm} 
    \begin{subfigure}[t]{.33\textwidth}
        \centering
        \includegraphics[width=\textwidth]{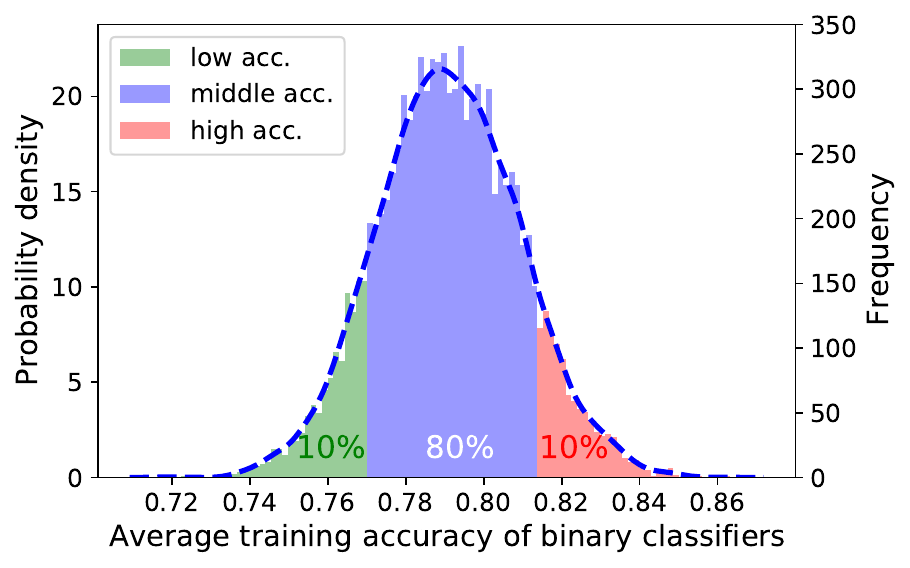}
        \caption{on the \textit{Digit} with 10 classes}
        \label{fig:accueach_each_digit}
    \end{subfigure} \hspace{-0.2cm}
    \begin{subfigure}[t]{.33\textwidth}
        \centering
        \includegraphics[width=\textwidth]{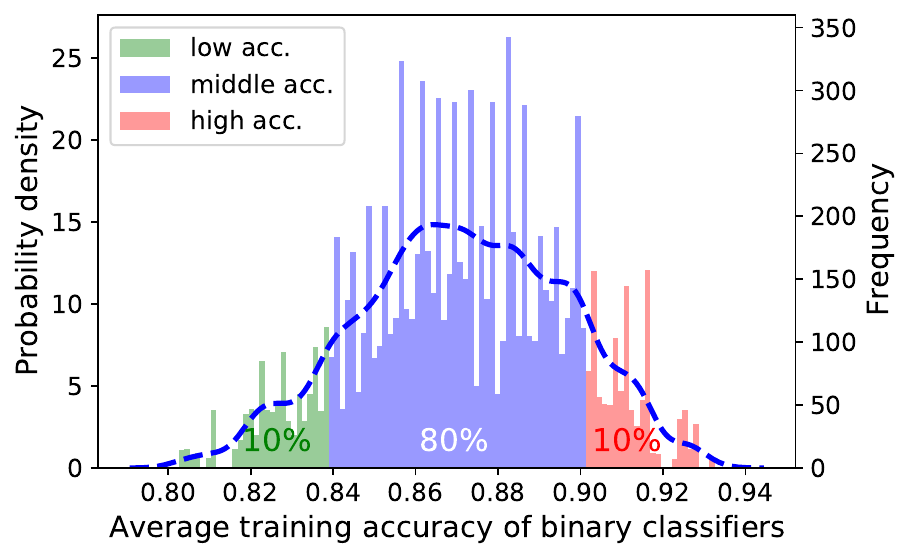}
        \caption{on the \textit{Ecoli.} with 8 classes}
        \label{fig:accueach_each_ecoli}
    \end{subfigure} 
    \caption{Distribution of all minimal ECOC-NEAT in terms of average classifiers training accuracy on the three datasets of \textit{Satellite}, \textit{Digit}, \textit{Ecoli.}. The frequency of the right vertical axis represents the number of ECOC.}
    \label{fig:accueach}
\end{figure*}

For the \textit{Satellite} dataset with $k=6$ classes, the minimal ECOC-NEAT needs a minimum of 3-bit codeword (three columns) to construct the ECOC. 
We random choose three columns from the 31 columns of the exhaustive ECOC to construct minimal ECOCs.
For an exhaustive ECOC with 31 columns, there are $\binom {31}{3} = 4495$ combinations, and 420 out of these 4495 combinations are available minimal ECOCs that satisfy both row and column conditions. 
We run all 420 minimal ECOC-NEAT on the \textit{Digit} dataset.
\autoref{fig:accueach} (a) shows the distribution of average training accuracy of binary classifiers (noted as $\overline{\mathcal{A}_b}$) in these 420 minimal ECOCs.
These 420 minimal ECOCs perform different qualities in terms of their average training accuracy of binary classifier from around $70\%$ to $90\%$.
We divide these 420 minimal ECOCs into the three-level performance of low, middle, high accuracy with the ratio of 10\%, 80\%, and 10\% respectively. 
These 10\% minimal ECOCs with high accuracy are the optimized minimal ECOCs.
The results indicate that different ECOCs perform significantly accuracy and the quality of ECOC is crucial for the high accuracy of binary classifiers.

Moreover, we randomly choose minimal ECOCs with low, middle, and high accuracy respectively.
Each minimal ECOC-NEAT evolves three binary classifiers (three columns in each minimal ECOC) with an evolution of total 3000 generations for multiclass classification, which results is shown in \autoref{tab:ecoc_choice_3datasets}.
The results indicate that the average training accuracy of binary classifiers significantly impacts the testing accuracy. 
The optimized minimal ECOC-NEAT performs a testing accuracy of $0.7735$ that is much higher than the low accuracy minimal ECOC-NEAT and the standard NEAT ($0.6377$ in \autoref{tab:result3dataset}) for 6-classes classification on the \textit{Satellite} dataset.
Conversely, the low accuracy minimal ECOC-NEAT perform a similar testing accuracy with the standard NEAT.

\begin{table}[!ht]
\centering \small
\renewcommand{\arraystretch}{1.0}
\setlength\tabcolsep{4pt} 
\caption{The performance of minimal ECOC-NEAT of different qualities on the three datasets of \textit{Satellite}, \textit{Digit}, and \textit{Ecoli}. $\overline{\mathcal{A}_b}$ represents the average training accuracy of binary classifiers.}
\begin{tabular}{l l c c c c c}
\toprule 
Dataset & \makecell[l]{minimal \\ ECOC} & \makecell[l]{Testing \\ accuracy} & Variance & \makecell[l]{Training \\ accuracy} & $\overline{\mathcal{A}_b}$ \\ \midrule
\multirow{3}{*}{Satellite} & middle- & 0.704  & 9.79$\times10^{-4}$ & 0.714 & 0.881 \\ 
& high-  & \underline{\textbf{\colorbox{lightgray!60}{0.774}}} & \textbf{\colorbox{lightgray!60}{3.55$\times10^{-4}$}} & \textbf{\colorbox{lightgray!60}{0.793}} & \textbf{\colorbox{lightgray!60}{0.915}} \\ 
& low- &  0.632 & 33.77$\times10^{-4}$  & 0.653 & 0.848 \\ \midrule
\multirow{3}{*}{Digit} & middle- & 0.535  & 72.01$\times10^{-4}$ & 0.614  &  0.865 \\ 
& high- & \underline{\textbf{\colorbox{lightgray!60}{0.647}}} & 31.44$\times10^{-4}$ & \textbf{\colorbox{lightgray!60}{0.733}}  & \textbf{\colorbox{lightgray!60}{0.912}}  \\ 
& low- & 0.483 & \textbf{\colorbox{lightgray!60}{6.83$\times10^{-4}$}} & 0.538 & 0.814 \\ \bottomrule
\multirow{3}{*}{Ecoli.} & middle- & 0.765  & $2.28\times10^{-4}$ & 0.816 &  0.922 \\ 
& high- & \underline{\textbf{\colorbox{lightgray!60}{0.818}}} & 1.86$\times10^{-4}$ & \textbf{\colorbox{lightgray!60}{0.867}}  & \textbf{\colorbox{lightgray!60}{0.953}}  \\ 
& low- &  0.678 & \textbf{\colorbox{lightgray!60}{1.76$\times10^{-4}$}}  & 0.740 &  0.870  \\ \bottomrule
\end{tabular}
\label{tab:ecoc_choice_3datasets} 
\end{table}

\begin{figure*}[!ht]
\centering
    \begin{subfigure}[t]{.32\textwidth}
        \centering
        \includegraphics[width=\textwidth]{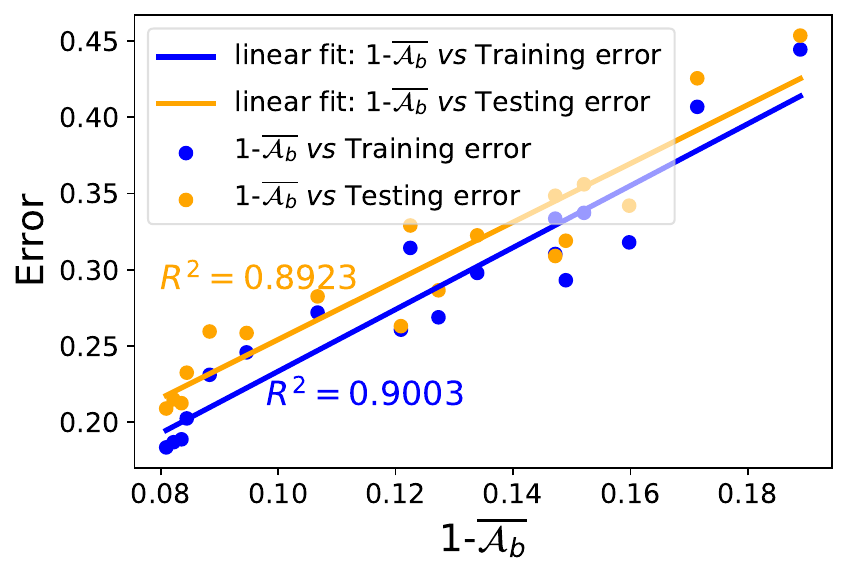}
        \caption{Satellite (6 classes)}
        \label{fig:linear_satellite}
    \end{subfigure} 
    \begin{subfigure}[t]{.32\textwidth}
        \centering
        \includegraphics[width=\textwidth]{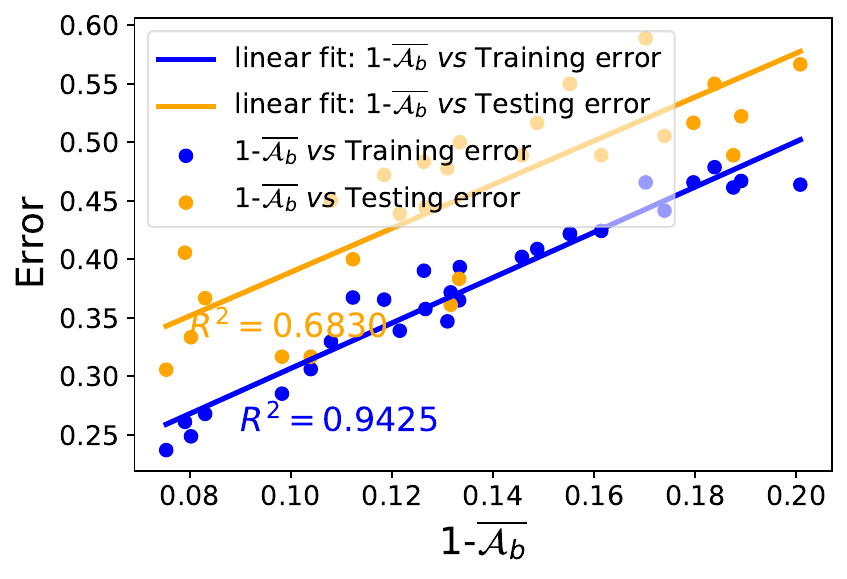}
        \caption{Digit (10 classes)}
        \label{fig:linear_digit}
    \end{subfigure} 
    \begin{subfigure}[t]{.32\textwidth}
        \centering
        \includegraphics[width=\textwidth]{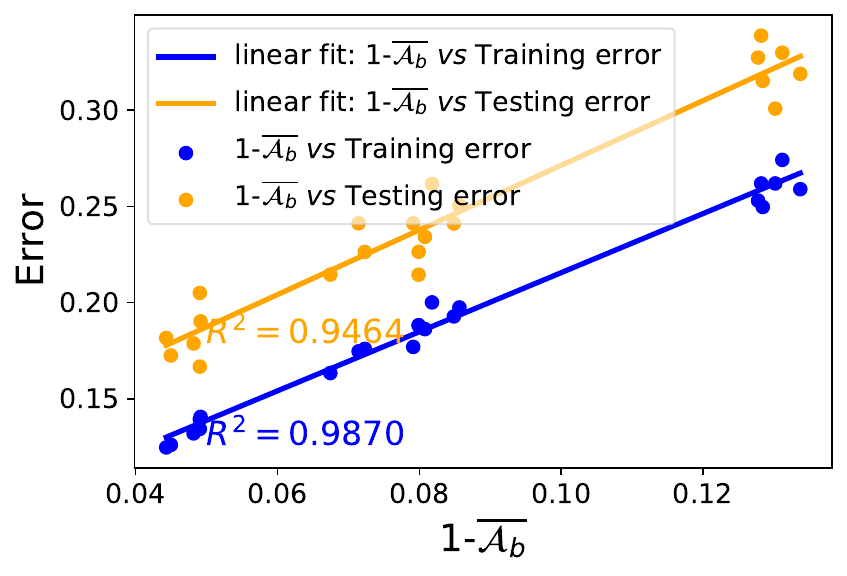}
        \caption{Ecoli. (8 classes)}
        \label{fig:linear_ecoli}
    \end{subfigure} 
    \caption{Training/Testing error and average training error of binary classifiers on the \textit{Satellite} problem. Distribution of all minimal ECOC-NEAT in terms of average classifiers training accuracy on the three datasets of \textit{Satellite}, \textit{Digit}, \textit{Ecoli.}. The frequency of the right vertical axis represents the number of ECOC. The lines are applied to fit the data, and $R^2$ is the goodness of fit.}
    \label{fig:linear}
\end{figure*}

Finally, we randomly choose 6 ECOCs from high, middle, low accuracy ECOCs respectively, that is 18 various ECOCs in total, to observe the relationship between their training/testing error and average training error of binary classifiers (1-$\overline{\mathcal{A}_b}$), as shown in \autoref{fig:linear} (a).
The lines are applied to fit the data, and indicate that the training/testing error is linear with the average training error of binary classifiers.
The optimized minimal ECOCs perform the results that are shown in the left-bottom points with low training/testing error and low average training error of binary classifiers.

\subsubsection{on the \textit{Digit} Dataset}

For the \textit{Digit} dataset with 10 classes, an exhaustive ECOC and a minimal ECOC consists of 511 base classifiers and four base classifiers respectively (as shown in \autoref{tab:result3dataset}). 
An exhaustive ECOC with 511 columns can be used to construct a large number of $\binom {511}{4} = 2,807,768,705$ 4-bit possible minimal ECOCs (4 columns) that is a huge amount of work and not necessary to be investigated. 
In this work, we random choose 10,000 minimal ECOCs to investigate the performance of various minimal ECOCs on the \textit{Digit} dataset.
The distribution of the average training accuracy of binary classifiers is shown in \autoref{fig:accueach} (b). 
Interestingly, the distribution looks like a normal distribution. 
We divide these minimal ECOCs into the three-level performance of low, middle, high accuracy with the ratio of 10\%, 80\%, and 10\% respectively.
As the standard NEAT with an evolution of 3000 generations, each classifier of these 511 binary classifiers is generated by an evolution of $\lceil 3000/511 \rceil \approx 6$ generations.
Theoretically and empirically, the average training accuracy of binary classifiers can be improved with a longer evolution than 6 generations, and thus lead to the higher accuracy for multiclass classification on the \textit{Digit} dataset.

We randomly choose minimal ECOCs from low, middle, high accuracy (in \autoref{fig:accueach} (b)) respectively.
These minimal ECOC-NEAT evolve binary classifiers with an evolution of $3000/4 = 750$ generations. 
The results of these minimal ECOC-NEAT on the \textit{Digit} dataset is shown in \autoref{tab:ecoc_choice_3datasets}. 
The high accuracy 4-bit ECOC-NEAT performs a remarkable testing accuracy that is comparable with the 10-bit mid-length ECOC-NEAT (a testing accuracy of 0.6506, see \autoref{tab:result3dataset}), and saves 60\% classifiers (from 10 to 4). 
The low accuracy ECOC-NEAT still perform a low testing accuracy of 0.4832 that is only a little superior to the standard NEAT.

We randomly choose 9 minimal ECOCs from low, middle, high accuracy respectively, that is 27 various ECOCs in total, to investigate the relationship between their training/testing error and average training error of binary classifiers (1-$\overline{\mathcal{A}_b}$), as shown in \autoref{fig:linear} (b). 
The lines are applied to fit the data, and indicate that the training/testing error is linear with the average training error of binary classifiers.
The 27 minimal ECOC-NEAT generate binary classifiers by an evolution of $3000/4 = 750$ generations and thus the binary classifiers performs higher average training accuracy (1 - average training error of binary classifiers) than the results in \autoref{fig:accueach} (b).

\paragraph{on the Ecoli. Dataset}

For the \textit{Ecoli.} dataset with 8 classes, an exhaustive ECOC-NEAT and a minimal ECOC-NEAT consists of 127 and 3 base classifiers (3-bit) respectively. 
An exhaustive ECOC can be used to construct a large number ($\binom {127}{3} = 333,375$) of minimal ECOCs. 
In this work, we randomly choose $10,000$ minimal ECOCs. 
The distribution of average training accuracy of binary classifiers is shown in \autoref{fig:accueach} (c).
We categorize these minimal ECOC-NEAT into three levels of high, middle, low average training accuracy of binary classifiers. 

Moreover, we randomly choose a minimal ECOC from low, middle, high accuracy respectively and run the minimal ECOC-NEAT to evolve binary classifiers with an evolution of 1000 (3000/3) generations. 
The results of the low, middle, high accuracy (optimized) minimal ECOC-NEAT on the \textit{Ecoli} dataset are shown in \autoref{tab:ecoc_choice_3datasets}. 
The high accuracy 3-bit minimal ECOC-NEAT performs a test accuracy near with 15-bit mid-length ECOC-NEAT. 
The low accuracy ECOC-NEAT performs a low test accuracy of 0.6782 that is even lower than that of the standard NEAT.

In addition, we randomly choose 7 minimal ECOCs from low, middle, high accuracy (optimized) minimal ECOCs (i.e., 21 various ECOCs in total) to validate the relationship between the quality of ECOCs and their training/testing error, as shown in \autoref{fig:linear} (c).
The lines that fit the results and indicate the linear relation between the quality of ECOCs and their training/testing error.

To summarize, we conclude that a high quality ECOC generally performs high testing accuracy. 
It is crucial to design a high quality ECOC for multiclass classification of neuroevolution approaches.

\subsection{Evolutionary Efficiency}
\label{sec:evolutionaryefficiency}

\begin{figure}[!ht]
\centering
\includegraphics[width=0.48\textwidth]{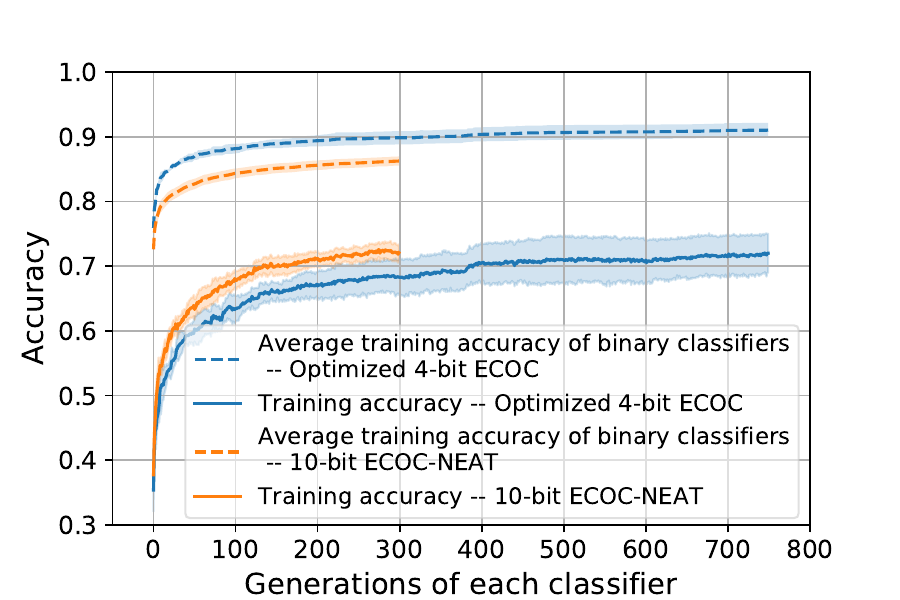} 
\caption{The training accuracy and average training accuracy of binary classifiers of 4-bit optimized minimal ECOC-NEAT and 10-bit mid-length ECOC-NEAT on the \textit{Digit} dataset. The lines and shadow represent the mean and 95\% confidence intervals for 10 repetitions.}
\label{fig:training_curves4b10}
\end{figure}

We observe the convergence of training accuracy and average training accuracy of binary classifiers during the evolution.
We randomly choose an optimized minimal ECOC-NEAT from \autoref{tab:ecoc_choice_3datasets} and a 10-bit ECOC-NEAT from \autoref{tab:result3dataset}, and run them 10 repetitions on the \textit{Digit} dataset.
The minimal ECOC-NEAT and 10-bit mid-length ECOC-NEAT generate binary classifiers with an evolution of 750 generations and 300 generations respectively.
The results are shown in \autoref{fig:training_curves4b10}.

The results show that the training accuracy performs a significant similar convergence process with the average training accuracy of binary classifiers. 
Both of them dramatically increase in the beginning and gradually converge to a stable value over generations. 
The high accuracy 4-bit minimal ECOC-NEAT performs a training accuracy of 72\% approximately, which performs even higher accuracy than the 10-bit mid-length ECOC-NEAT with a training accuracy of 71\%.

\begin{figure}[!ht] 
\centering
\includegraphics[width=0.48\textwidth]{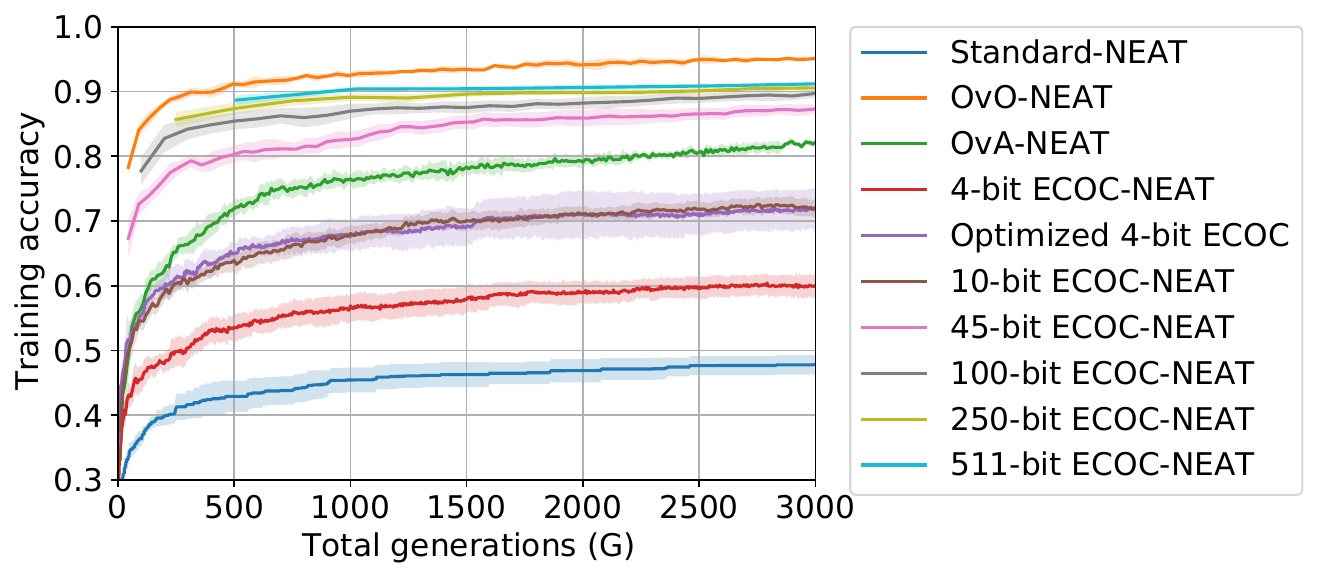} 
\caption{Training accuracy of the standard NEAT and the NEAT with class binarization techniques over generations on the \textit{Digit} dataset for 10 classes classification.} 
\label{fig:training_curves}
\end{figure}

Moreover, we compare the training accuracy of the standard NEAT and NEAT with class binarization techniques during the evolution, as shown in \autoref{fig:training_curves}.
The number of generations for each evolution of ECOC-NEAT is $\mathcal{G}/\mathcal{N}$ which is different for various ECOC-NEAT.
To compare various ECOC-NEAT in the same scale, we apply proportional scaling to match an identical x-axis.
For example, 10-bit mid-length ECOC-NEAT with the evolution of 300 generations for each binary classifier in \autoref{fig:training_curves4b10} is scaled 10 times in \autoref{fig:training_curves}. 

The results show that NEAT with class binarization techniques perform significantly better in terms of accuracy than the standard NEAT for multiclass classification. 
OvO-NEAT, exhaustive ECOC-NEAT, mid-length ECOC-NEAT (including 250-bit, 100-bit, 45-bit ECOC-NEAT) perform remarkable training accuracy.
The NEAT with large size ECOC (e.g., exhaustive ECOC-NEAT, OvO-NEAT) generally performs better than the NEAT with small size ECOC (e.g., 4-bit ECOC-NEAT).
Compared to the normal 4-bit ECOC-NEAT with a training accuracy of 60\% approximately, the optimized 4-bit ECOC-NEAT perform an efficient multiclass classification with a training accuracy of 72\% approximately.
Moreover, the optimized 4-bit ECOC-NEAT performs significantly similar evolution process (the purple line)  with the 10-bit ECOC-NEAT (the brown line). 
The results demonstrate that the size and quality of ECOC are crucial for the multiclass classification performance of ECOC-NEAT.

\subsection{Robustness}
\label{sec:robustness}

\begin{figure}[ht!] 
\centering 
\includegraphics[width=0.4\textwidth]{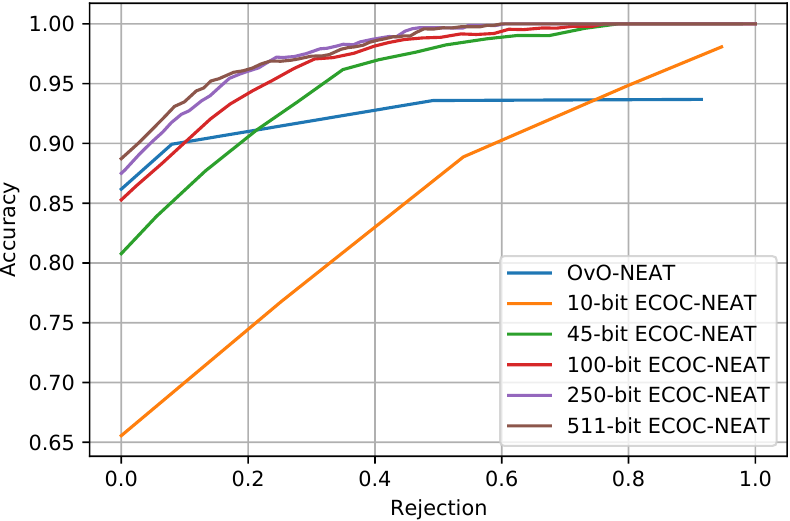} 
\caption{Accuracy rejection curves of OvO-NEAT and other ECOC-NEAT on the \textit{Digit} dataset.} 
\label{fig:rejection_curve} 
\end{figure}

Robustness is an important measurement for the evaluation of multiclass classification.
The ECOC-NEAT usually performs a remarkable ability to correct errors for multiclass classification. 
Gunjan Verma and Ananthram Swami applied ECOC to improve the adversarial robustness of deep neural networks \cite{verma2019error}. 
Although OvO-NEAT performs outstanding for multiclass classification, the robustness of OvO-NEAT against errors is insufficient compared to ECOC-NEAT. 
In this work, we apply the measure of Accuracy-Rejection curve to analyse the robustness of the NEAT with class binarization techniques.
\autoref{fig:rejection_curve} shows the accuracy-rejection curve of OvO-NEAT and other ECOC-NEAT. 

The large size ECOCs perform better than the small size ECOCs no matter whether the rejection rates are low or high.
Large size ECOC-NEAT always outperforms OvO-NEAT, which means they have consistently stronger robustness against errors than OvO-NEAT. 
Comparing the small size of 10-bit ECOC-NEAT with OvO-NEAT, there is an intersection between two lines. 
The lines of the large size ECOC intersect the line of OvO-NEAT at the small values of rejections. 
From a rejection rate of the intersection onwards, ECOC-NEAT outperforms OvO-NEAT. 
For example, at rejection rates greater than 80\%, even 10-bit ECOC-NEAT outperforms OvO-NEAT, which means 10-bit ECOC-NEAT gives 20\% of the test samples pretty convincing predictions (with testing accuracy of 95\%). Briefly, ECOC-NEAT has strong robustness against errors, especially with long codes. By contrast, the robustness of OvO-NEAT seems weak.

\begin{figure}[!ht] 
\centering 
\includegraphics[width=0.42\textwidth]{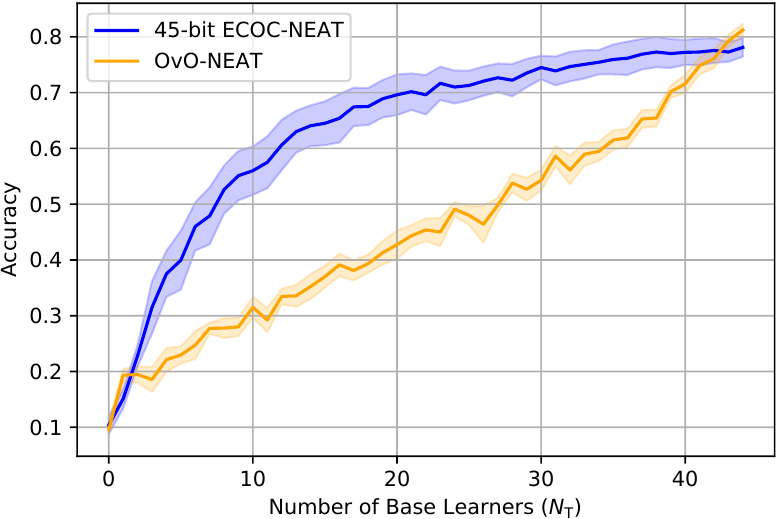} 
\caption{Testing accuracy of ECOC-NEAT and OvO-NEAT with various number of base classifiers. The experiments are repeated 10 times and take an average of the testing accuracy. The shadow represents 95\% confidence intervals.}
\label{fig:ovoecoc} 
\end{figure}

ECOC-NEAT performs strong robustness that its base classifiers are complement each other when the number of base classifiers decreases. 
In this work, we investigate the robustness of the performance of ECOC-NEAT and OvO-NEAT when their number of base classifiers decreases.
The results are shown in \autoref{fig:ovoecoc}, where the size of ECOC and OvO decrease from 45-bit (45 base classifiers) to 1-bit (one classifier).
We randomly choose base classifiers from 45-bit ECOC-NEAT and OvO-NEAT to construct various size ECOC-NEAT and OvO-NEAT with ten repetitions.
The results show that the testing accuracy of OvO-NEAT declines almost linearly as the number of base classifiers decreases. 
However, the accuracy of ECOC-NEAT decreases slightly as the number of base classifiers decreases.
In particular, the accuracy of ECOC-NEAT hardly decreases when ECOC-NEAT is with a little fewer base classifiers, e.g., 40-bit ECOC-NEAT. 
The ECOC-NEAT with 22 base classifiers, that is half of 45 base classifiers, still obtains a testing accuracy of approximately $70\%$ that dropping by $12\%$ from the testing accuracy of 45-bit ECOC-NEAT ($82\%$).
However, OvO-NEAT with 22 base classifiers performs $45\%$ testing accuracy that dropping by $41\%$ from the testing accuracy of 45-bit OvO-NEAT ($86\%$).
This finding illustrates that ECOC-NEAT performs better robustness than OvO-NEAT when they ensemble fewer base classifiers for multiclass classification.

OvO is a decent class binarization technique for multiclass classification with high accuracy, low variance, and efficient training process \cite{galar2011overview,mcdonnell2018divide}, but requires too many classifiers ($O(k^2)$).
The large size ECOC usually performs high accuracy, low variance, and strong robustness \cite{garcia2011empirical,kong1995error}.
An optimized minimal ECOC significantly outperforms a normal constructed ECOC \cite{bautista2012minimal}.

In summary, we recommend OvO-NEAT and ECOC-NEAT with a great number of binary classifiers (e.g. mid-length ECOC-NEAT, or exhaustive ECOC-NEAT with moderate classes) for the tasks when a considerable number of generations is allowed.
For the tasks that only limited generations are allowed, we recommend optimized ECOC-NEAT with a small number of binary classifiers.

\section{Discussions and Future work}
\label{sec:discussionandfuture}

\subsection{Discussions}
\label{subsec:dis}
 
In this section, we analyse the classification performance of these methods on different classes and the network complexity of base classifiers.

\subsubsection{Behavior Analysis}
\label{subsubsec:behavioranalysis}
\begin{figure}[!ht]
\centering 
\includegraphics[width=0.48\textwidth]{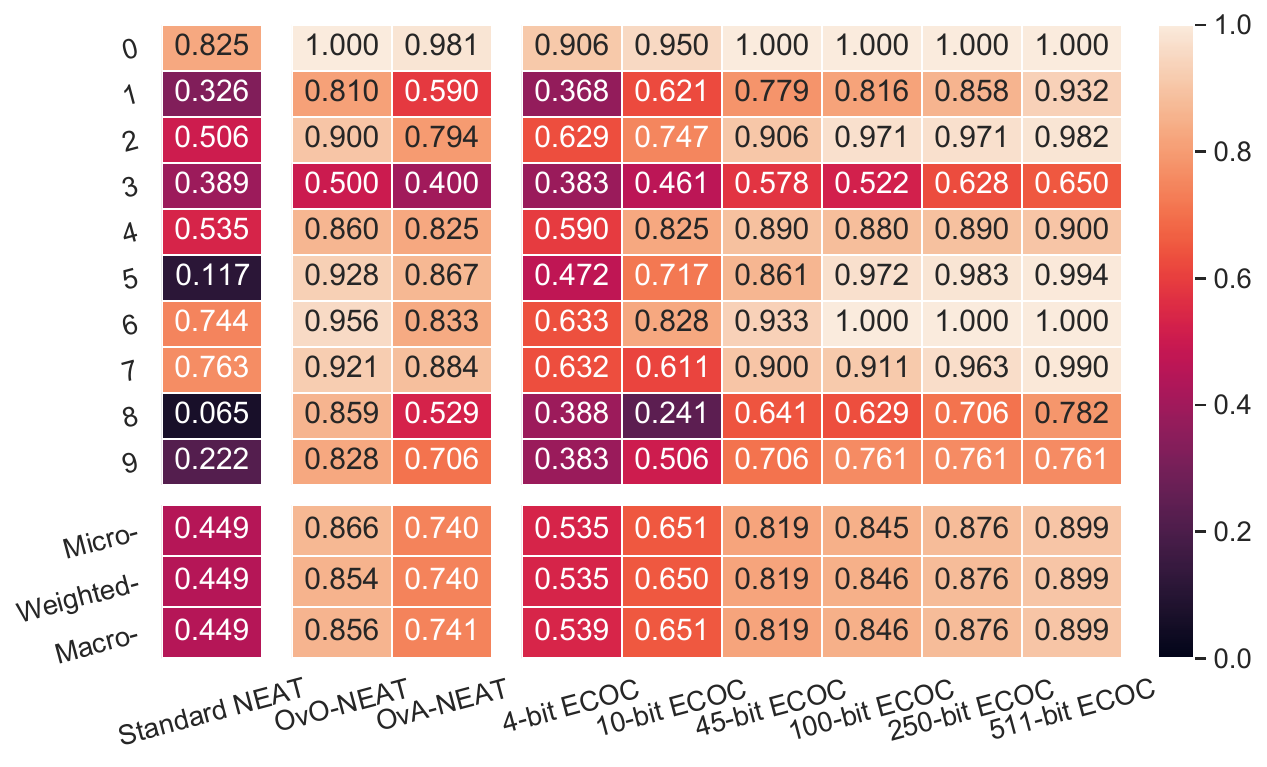} 
\caption{Precision heatmap of these methods on the \textit{Digit} dataset. Rows from 0 to 9 are precision on the digit class from "0" to "9". Rows 10, 11, 12 present micro-averaging precision, weighted-averaging precision, and macro-averaging precision, respectively. Columns represent various methods.
}
\label{fig:precision_heatmap} 
\end{figure}

\begin{figure}[!ht]
\centering 
\includegraphics[width=0.48\textwidth]{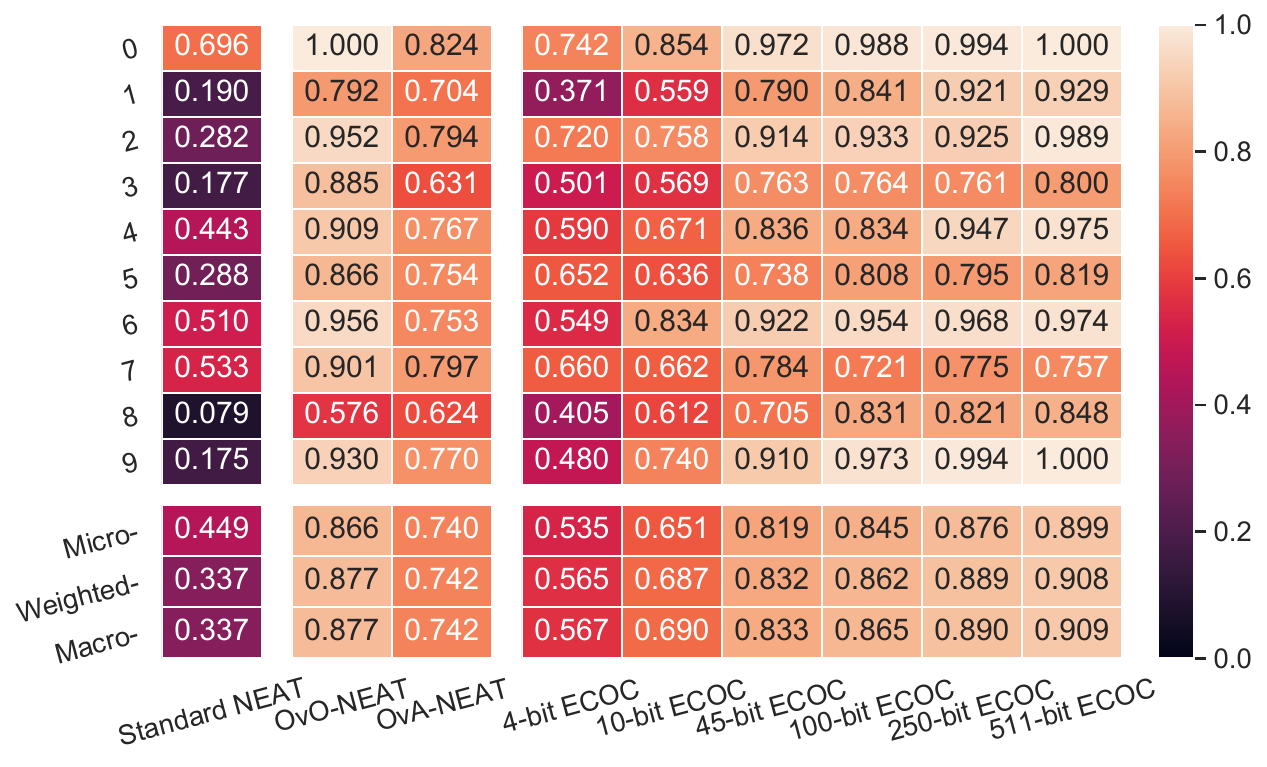} 
\caption{Recall heatmap of different methods on the \textit{Digit} dataset. Rows 0 to 9 present the recall of digit class "0" to "9". Rows 10 to 12 present micro-averaging recall, weighted-averaging recall, and macro-averaging recall, respectively. Columns represent different methods.} 
\label{fig:recall_heatmap} 
\end{figure}

\begin{figure}[!ht] 
\centering 
\includegraphics[width=0.48\textwidth]{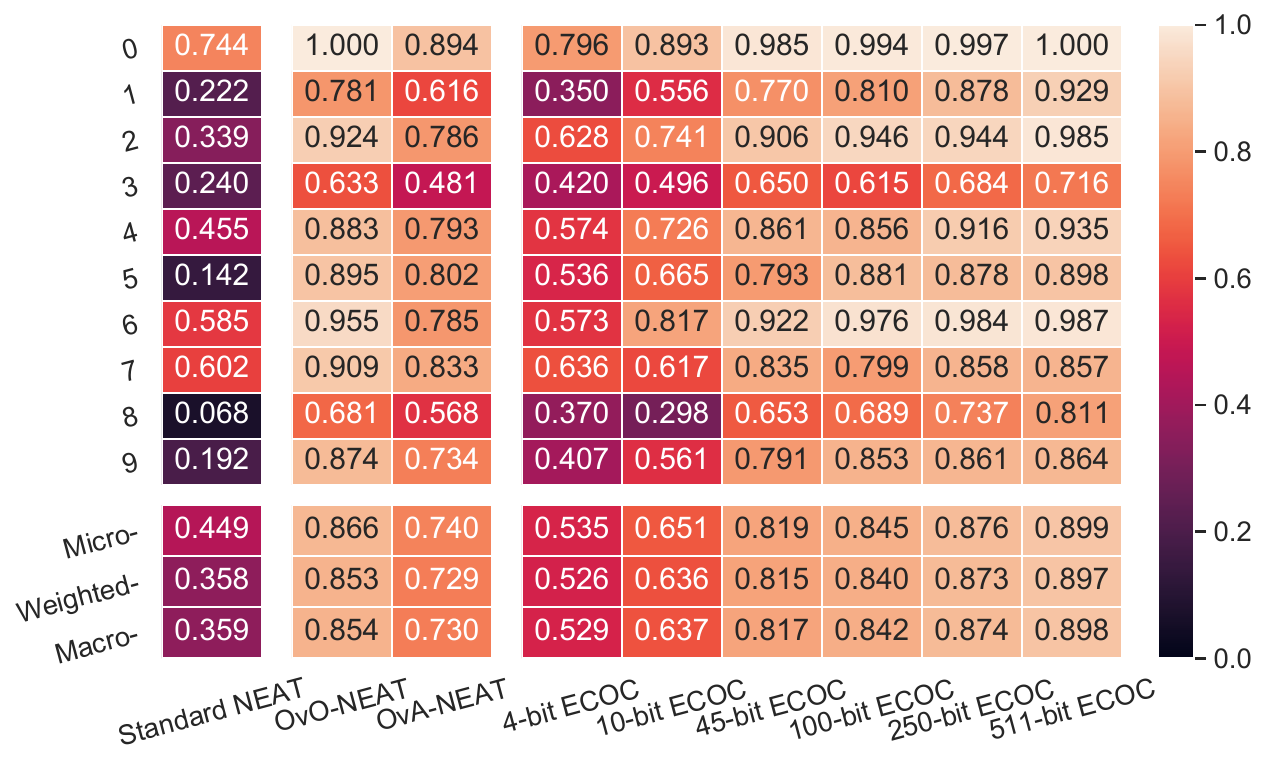} 
\caption{F1-score Heatmap of different multiclass classification methods. Rows 0 to 9 present the F1-score of digit class from "0" to "9". Rows 10 to 12 present micro-averaging, weighted-averaging, and macro-averaging F1-score, respectively. Columns represent different methods.}
\label{fig:f1_heatmap} 
\end{figure}

We observe the classification performance on each class of these methods by analyzing the results of the standard NEAT and the NEAT with class binarization techniques on the \textit{Digit} dataset \footnote{It does not need to analyze the results on all three datasets}.
We apply the widely used metrics of precision, recall, and F1-score to evaluate the classification on each class of these methods.
Moreover, we adopt popular averaging methods for precision, recall, and F1-score, resulting in a set of different average scores (macro-averaging, weighted-averaging, micro-averaging), see more details of these averaging methods in \cite{grandini2020metrics}.
We conduct experiments for ten repetitions and take an average of the results. 
The heatmaps of the precision, recall, and F1-score of these methods are visualized in \autoref{fig:precision_heatmap}, \autoref{fig:recall_heatmap}, \autoref{fig:f1_heatmap}, respectively.

The classification precision on each class of the \textit{Digit} dataset from "0" to "9" is shown in the heatmap of \autoref{fig:precision_heatmap}.
The results show that the difficulty of classifications on different digits is diverse.
Specifically, the digit "0" is predicted by all these methods with high accuracy of more than 90\%.
All these methods perform low testing accuracy on the digit "3" and "8".
The other digits are classified with diverse accuracies that are basically desired.
The larger size ECOC-NEAT generally performs higher precision than the small size ECOC-NEAT.
For example, a micro-averaging precision of 0.5350 for 4-bit ECOC-NEAT increases to 0.8189 for 45-bit ECOC-NEAT.
All ECOC-NEAT including the small size 4-bit ECOC-NEAT outperform the standard NEAT.
The precision of the standard NEAT once again verifies its low performance for multiclass classification. 
Exceptionally, the standard NEAT predict the digit "0" with a decent accuracy, which verifies that the digit "0" is distinctly predicted.

\autoref{fig:recall_heatmap} shows the recall heatmap of different methods for classifying the digit class "0" to "9". 
The recall heatmap shows consistent results with the precision heatmap. 
For example, the recall of digit classes "3" and "8" are usually the low for all these methods.

F1-score is the harmonic mean of precision and recall to evaluate model performance comprehensively, which conveys a balance between precision and recall.
The F1-score of different methods on the \textit{Digit} dataset is shown in \autoref{fig:f1_heatmap}.
The recall heatmap shows consistent results with the precision and recall heatmaps.

\begin{figure}[!ht] 
\centering 
\includegraphics[width=0.43\textwidth]{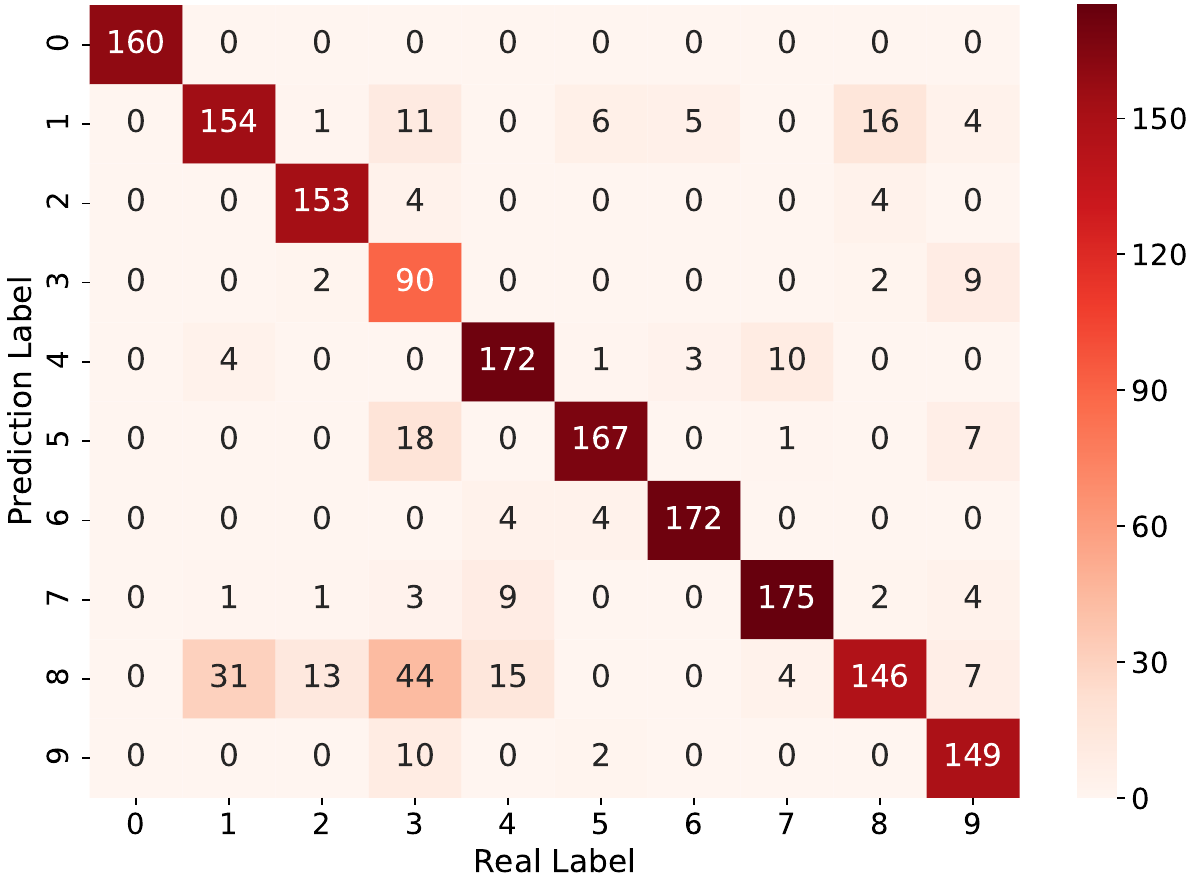} 
\caption{The heatmap of predicted label by OvO-NEAT and real label on the \textit{Digit} dataset.} 
\label{fig:ovocm} 
\end{figure}

It is worth noticing that OvO-NEAT performs a high precision on the digit "8" but a low precision on the digit "3" in \autoref{fig:precision_heatmap}. 
By contrast, its recall on the digit "8" is lower compared to the digit "3" in \autoref{fig:recall_heatmap}. 
We suppose that there are recognition errors between these two categories, and therefore observe the predicted label of OvO-NEAT and real label to verify this hypothesis, as shown in \autoref{fig:ovocm}.
The results show that OvO-NEAT often incorrectly predicts the digit "3" as "8".
Specifically, 44 digits of "3" are incorrectly predicted as the digit "8".
This explains that these methods perform low testing accuracy on the digit "3" and "8".
Intuitively, the digit "3" and "8" have similar shapes, and they are even incorrectly recognized by human.

In summary, the three heatmaps of precision, recall, and F1-score reveal consistent conclusions that 1) NEAT with class binarization techniques, particularly ECOC-NEAT and OvO-NEAT, outperform the standard NEAT for multiclass classification, 2) the large ECOC-NEAT generally performs high precision, recall, and F1-score, 3) NEAT (including the standard NEAT, OvO-NEAT, ECOC-NEAT) techniques perform diverse on different classes and large size ECOC-NEAT perform robust for the classification with different classes.

\subsubsection{Network Complexity}
\label{subsubsec:networkcomplexity}

Network complexity offers an insight into the analysis of the mechanisms of NEAT with class binarization techniques for multiclass classification. 
We investigate how the number of nodes and connections influence classification performance. 
\autoref{tab:complexity_multi} shows the network complexity of generated classifiers by different NEAT-based methods for a different number of classes on the \textit{Digit} dataset.
These experiments are repeated ten times. 
The network complexity on the \textit{Satellite} and \textit{Ecoli.} dataset are presented in \autoref{tab:nodessatellite} and \autoref{tab:nodesecoli}.
We observe the average total number of nodes and connections of all base classifiers over ten repetitions, and the average number of nodes and connections of each base classifier (the value in the bracket). 
For example, the exhaustive ECOC-NEAT generate three base classifiers with an average total number of 107 nodes and 286 connections for 3 classes, and an average number of 36 nodes and 95 connections for each base classifier over ten repetitions.

\begin{table*}[!ht]
\centering
\small
\setlength{\tabcolsep}{3pt}
\caption{Network complexity of generated classifiers by different NEAT-based methods for different number of classes on the \textit{Digit} dataset. The value and the the value in the bracket are the average total number of all base classifiers and the average number of each base classifier over ten repetitions, respectively.}
\begin{tabular}{c c c c c c c c c c}
\toprule
&  & \multicolumn{8}{c}{Number of Classes} \\ \cline{3-10}
&  & 3 & 4 & 5 & 6 & 7 & 8 & 9 & 10 \\ \midrule
\multirowcell{4}{Standard\\NEAT} & \# Classifiers & 1 & 1 & 1 & 1 & 1 & 1 & 1 & 1 \\
& Generations & 3000(3000) & 3000(3000) & 3000(3000) & 3000(3000) & 3000(3000) & 3000(3000) & 3000(3000) & 3000(3000) \\
& Nodes & 43(43) & 68(68) & 61(61) & 55(55) & 60(60) & 63(63) & 68(68) & 72(72) \\
& Connections & 150(150) & 308(308) & 211(211) & 172(172) & 133(133) & 121(121) & 132(132) & 177(177) \\ \cline{2-10}
\multirowcell{4}{OvO-NEAT} & \# Classifiers & 3 & 6 & 10 & 15 & 21 & 28 & 36 & 45 \\
& Generations & 3000(1000) & 3000(500) & 3000(300) & 3000(200) & 3003(143) & 2996(107) & 2988(83) & 3015(67) \\
& Nodes & 81(27) & 159(27) & 249(25) & 396(26) & 506(24) & 657(23) & 857(24) & 1039(23) \\
& Connections & 205(68) & 282(47) & 505(51) & 658(44) & 744(35) & 916(33) & 1177(33) & 1387(31) \\ \cline{2-10}
\multirow{4}{*}{OvA-NEAT} & \# Classifiers & 3 & 4 & 5 & 6 & 7 & 8 & 9 & 10 \\
& Generations & 3000(1000) & 3000(750) & 3000(600) & 3000(500) & 3003(429) & 3000(375) & 2997(333) &  3000(300) \\
& Nodes & 109(36) & 144(36) & 186(37) & 221(37) & 260(37) & 303(38) & 325(36) & 355(36) \\
& Connections & 341(114) & 464(116) & 661(132) & 649(108) & 598(85) & 967(121) & 836(93) & 931(93) \\ \cline{2-10}
\multirowcell{4}{Minimal\\ECOC-NEAT} & \# Classifiers & 2 & 2 & 3 & 3 & 3 & 3 & 4 & 4 \\
& Generations & 3000(1500) & 3000(1500) & 3000(1000) & 3000(1000) & 3000(1000) & 3000(1000) & 3000(750) & 3000(750) \\
& Nodes & 77(39) & 98(49) & 132(44) & 142(47) & 139(46) & 120(40) & 166(42) & 174(44) \\
& Connections & 207(104) & 492(246) & 475(158) & 479(160) & 471(157) & 493(164) & 518(130) & 540(135) \\ \cline{2-10}
\multirowcell{4}{Mid-length\\ECOC-NEAT} & \# Classifiers & 3 & 7 & 15 & 26 & 29 & 30 & 32 & 34 \\
& Generations & 3000(1000) & 3003(429) & 3000(200) & 2990(115) & 2987(103) & 3000(100) & 3008(94) & 2992(88) \\
& Nodes & 107(36) & 266(38) & 477(32) & 739(28) & 796(27) & 831(28) & 848(27) & 881(26) \\
& Connections & 286(95) & 783(112) & 1012(67) & 1300(50) & 1450(50) & 1347(45) & 1353(42) & 1416(42) \\ \cline{2-10}
\multirowcell{4}{Exhaustive\\ECOC-NEAT} & \# Classifiers & 3 & 7 & 15 & 31 & 63 & 127 & 255 & 511 \\
& Generations & 3000(1000) & 3003(429) & 3000(200) & 3007(97) & 3024(48) & 3048(24) & 3060(12) & 3066(6) \\
& Nodes & 107(36) & 266(38) & 477(32) & 836(27) & 1446(23) & 2988(24) & 4740(19) & 8711(17) \\
& Connections & 286(95) & 783(112) & 1012(67) & 1317(42) & 1903(30) & 3060(24) & 5060(20) & 8688(17) \\ \bottomrule
\end{tabular}
\label{tab:complexity_multi}
\end{table*}

As the number of classes increases, it is reasonable to generate a complex neural network with more nodes and connections for more complicated patterns. 
However, the results show that the standard NEAT struggles to generate the neural networks with augmented nodes and connections as the number of classes increases, which basically causes its dramatic multiclass classification degradation. 
We hypothesis that the standard NEAT tend to eliminate the evolved neural networks with more nodes and connections during the evolution. 
In contrast, NEAT with class binarization techniques tend to generate neural networks with more nodes and connections as the number of classes increases for remarkable multiclass classification.
Although ECOC-NEAT often generates the base classifiers with fewer and fewer nodes and connections as the number of classes increases, the increasing number of binary classifiers leads to the increasing total number of nodes and connections that contribute to the remarkable performance of multiclass classification.
For example, the base classifier evolved by the exhaustive ECOC-NEAT for 3 classes has an average of 36 nodes and 95 connections, but that for 10 classes has an average of only 17 nodes and 17 connections.
However, the total nodes and connections increase from 107 and 286 to 8711 and 8688 respectively for 10 classes classification.

\begin{table}[ht!]
\centering
\footnotesize
\setlength{\tabcolsep}{3pt}
\caption{Network complexity of generated classifiers by different NEAT-based methods on the \textit{Satellite} dataset.}
\begin{tabular}{c c c c c}
\toprule
Method & Generations & \#Classifiers & Nodes & Connections \\ \midrule
Standard NEAT &  3000 (3000) & 1 & 44\textbf{\colorbox{lightgray!60}{(44)}} & 156 \textbf{\colorbox{lightgray!60}{(156)}}\\ 
OvO-NEAT & 3000 (200)  & 15 & 325 (22)  & 1095 (73) \\ 
OvA-NEAT & 3000 (500)  & 6  & 151 (25) & 975 (163)\\ 
Minimal ECOC & 3000 (1000) & 3  & 82 (27) & 566 (189)\\ 
6-bit ECOC  & 3000 (500)  & 6  & 145 (24) & 775 (129)\\
10-bit ECOC  & 3000 (300)  & 10 & 226 (23) & 831 (83)\\
15-bit ECOC  & 3000 (200)  & 15 & 303 (20) & 1076 (72)\\
20-bit ECOC  & 3000 (150)  & 20 & 397 (20) & 1119 (56)\\
Exhaustive ECOC  & 3007 (97)  & \textbf{\colorbox{lightgray!60}{31}} & \textbf{\colorbox{lightgray!60}{529}} (17) &  \textbf{\colorbox{lightgray!60}{1252}} (40)\\ \bottomrule
\end{tabular}
\label{tab:nodessatellite} 
\end{table}

\begin{table}[ht!]
\centering \footnotesize
\setlength{\tabcolsep}{3pt}
\caption{Network complexity of generated classifiers by different NEAT-based methods on the \textit{Ecoli.} dataset.} 
\begin{tabular}{c c c c c}
\toprule
Method & Generations & \#Classifiers & Nodes & Connections \\ \midrule
Standard NEAT &  3000 (3000) & 1 &  29 \textbf{\colorbox{lightgray!60}{(29)}}   &  262 \textbf{\colorbox{lightgray!60}{(262)}}  \\ 
\specialrule{0em}{1pt}{1pt}
OvO-NEAT & 2996 (107)  & 28 & 170 (6) & 254 (9)  \\ 
OvA-NEAT & 3000 (375) & 8 & 89 (11) & 323 (40) \\ 
\specialrule{0em}{1pt}{1pt}
Minimal ECOC & 3000 (1000) & 3 & 44 (15) & 272 (91)  \\ 
8-bit ECOC   & 3000 (375) & 8 & 106 (13) & 457 (57)  \\ 
15-bit ECOC  & 3025 (200) & 15 & 183 (12) & 634 (42)  \\  
28-bit ECOC  & 2996 (107) & 28 & 309 (11) & 880 (31)  \\ 
40-bit ECOC  & 3000 (75) & 40 & 407 (10) & 1007 (25)  \\ 
60-bit ECOC  & 3000 (50) & 60 & 556 (9) & 1202 (20)  \\ 
Exhaustive ECOC  & 3048 (24) & \textbf{\colorbox{lightgray!60}{127}} & \textbf{\colorbox{lightgray!60}{964}}(8) & \textbf{\colorbox{lightgray!60}{1321}}(10) \\ \bottomrule
\end{tabular}
\label{tab:nodesecoli} 
\end{table}

\subsection{Future Work}
\label{subsec:futurework}
Although this work investigates the different class binarization techniques, there are still multiple open issues and possible future work that may provide new insights into ECOC-NEAT.
First, the ECOC-NEAT needs to train a lot of binary classifiers, which generally takes a lot of training time.
Second, the hamming distance for matching the predicted codeword and ECOC codewords is a basic matching strategy that needs to be improved.  
Third, the ECOC still needs to be improved with different code design.  
We would like to improve the performance of ECOC-NEAT from the aspects of:
\begin{itemize}
    \item {using sparse codes (i.e., $\mathbb{M} \in \{1, -1, 0\}$) instead of dense codes (i.e., $\mathbb{M} \in \{1, -1\}$), which are beneficial to efficient training \cite{allwein2000reducing}.}
    \item {using other decoding strategies like loss-based decoding instead of hamming distance to match the codewords of ECOC. Loss-based decoding generally contributes to good performance because of the "confidence" information \cite{allwein2000reducing}.}
    \item {applying low-density parity-check code to design the optimized ECOC.}
\end{itemize}

\section{Conclusion}
\label{sec:conclusion}

This work investigates class binarization techniques in neuroevolution and proposes the ECOC-NEAT method that applies ECOC to the neuroevolution algorithm of NEAT for multiclass classification. 
We investigate 1) the performance of NEAT with different class binarization techniques for multiclass classification from multiclass degradation, accuracy, training efficiency, and robustness on three popular datasets,
2) the performance of ECOC-NEAT with different size and quality of ECOC.
The results show that ECOC-NEAT offers various benefits compared to the standard NEAT and NEAT with other class binarization techniques for multiclass classification.
Large size ECOCs and optimized ECOCs generally contribute to better performance for multiclass classification.
ECOC-NEAT shows significant benefits in a flexible number of binary classifiers and strong robustness.
In future, ECOC-NEAT can be extended to other applications such as image classification and computer vision.
Moreover, ECOC can be applied to different neuroevolution algorithms for multiclass classification.

\section*{Code and Data Availability}
The code and data for this work are available at \url{https://github.com/lafengxiaoyu/NEAT-ensembles}

\section*{CRediT authorship contribution statement}
\textbf{Gongjin Lan:} Conceptualization, Methodology, Validation, Visualization, Investigation, Writing - original draft, Writing - review \& editing.
\textbf{Zhenyu Gao:} Conceptualization, Methodology, Coding and Validation, Visualization, Investigation, Writing - original draft, Writing - review \& editing.
\textbf{Lingyao Tong:} Writing - review \& editing.
\textbf{Ting Liu:} Writing - review \& editing.

\section*{Declaration of Competing Interest}
The authors declare that they have no known competing financial interests or personal relationships that could have appeared to influence the work reported in this paper.

\section*{Acknowledgment}

This work is partially supported by the Guangdong Natural Science Funds for Young Scholar (No: 2021A1515110641).





\bibliographystyle{IEEEtran}
\bibliography{bibliography}

\begin{thebibliography}{10}
\providecommand{\url}[1]{#1}
\csname url@samestyle\endcsname
\providecommand{\newblock}{\relax}
\providecommand{\bibinfo}[2]{#2}
\providecommand{\BIBentrySTDinterwordspacing}{\spaceskip=0pt\relax}
\providecommand{\BIBentryALTinterwordstretchfactor}{4}
\providecommand{\BIBentryALTinterwordspacing}{\spaceskip=\fontdimen2\font plus
\BIBentryALTinterwordstretchfactor\fontdimen3\font minus
  \fontdimen4\font\relax}
\providecommand{\BIBforeignlanguage}[2]{{%
\expandafter\ifx\csname l@#1\endcsname\relax
\typeout{** WARNING: IEEEtran.bst: No hyphenation pattern has been}%
\typeout{** loaded for the language `#1'. Using the pattern for}%
\typeout{** the default language instead.}%
\else
\language=\csname l@#1\endcsname
\fi
#2}}
\providecommand{\BIBdecl}{\relax}
\BIBdecl

\bibitem{aly2005survey}
M.~Aly, ``Survey on multiclass classification methods,'' \emph{Neural Netw},
  vol.~19, pp. 1--9, 2005.

\bibitem{lorena2008review}
A.~C. Lorena, A.~C. De~Carvalho, and J.~M. Gama, ``A review on the combination
  of binary classifiers in multiclass problems,'' \emph{Artificial Intelligence
  Review}, vol.~30, no.~1, pp. 19--37, 2008.

\bibitem{galar2011overview}
M.~Galar, A.~Fern{\'a}ndez, E.~Barrenechea, H.~Bustince, and F.~Herrera, ``An
  overview of ensemble methods for binary classifiers in multi-class problems:
  Experimental study on one-vs-one and one-vs-all schemes,'' \emph{Pattern
  Recognition}, vol.~44, no.~8, pp. 1761--1776, 2011.

\bibitem{zhou2008data}
J.~Zhou, H.~Peng, and C.~Y. Suen, ``Data-driven decomposition for multi-class
  classification,'' \emph{Pattern Recognition}, vol.~41, no.~1, pp. 67--76,
  2008.

\bibitem{furnkranz2002round}
J.~F{\"u}rnkranz, ``Round robin classification,'' \emph{Journal of Machine
  Learning Research}, vol.~2, no. Mar, pp. 721--747, 2002.

\bibitem{mcdonnell2018divide}
T.~McDonnell, S.~Andoni, E.~Bonab, S.~Cheng, J.-H. Choi, J.~Goode, K.~Moore,
  G.~Sellers, and J.~Schrum, ``Divide and conquer: neuroevolution for
  multiclass classification,'' in \emph{Proceedings of the Genetic and
  Evolutionary Computation Conference}, 2018, pp. 474--481.

\bibitem{floreano2008neuroevolution}
D.~Floreano, P.~D{\"u}rr, and C.~Mattiussi, ``Neuroevolution: from
  architectures to learning,'' \emph{Evolutionary intelligence}, vol.~1, no.~1,
  pp. 47--62, 2008.

\bibitem{chen2006neuroevolution}
L.~Chen and D.~Alahakoon, ``Neuroevolution of augmenting topologies with
  learning for data classification,'' in \emph{2006 International Conference on
  Information and Automation}.\hskip 1em plus 0.5em minus 0.4em\relax IEEE,
  2006, pp. 367--371.

\bibitem{lan2019evolving}
G.~Lan, L.~De~Vries, and S.~Wang, ``Evolving efficient deep neural networks for
  real-time object recognition,'' in \emph{2019 IEEE Symposium Series on
  Computational Intelligence (SSCI)}.\hskip 1em plus 0.5em minus 0.4em\relax
  IEEE, 2019, pp. 2571--2578.

\bibitem{hagg2017evolving}
A.~Hagg, M.~Mensing, and A.~Asteroth, ``Evolving parsimonious networks by
  mixing activation functions,'' in \emph{Proceedings of the Genetic and
  Evolutionary Computation Conference}, 2017, pp. 425--432.

\bibitem{lan2021learning}
G.~Lan, M.~De~Carlo, F.~van Diggelen, J.~M. Tomczak, D.~M. Roijers, and A.~E.
  Eiben, ``Learning directed locomotion in modular robots with evolvable
  morphologies,'' \emph{Applied Soft Computing}, vol. 111, p. 107688, 2021.

\bibitem{gao2021neat}
Z.~Gao and G.~Lan, ``A neat-based multiclass classification method with class
  binarization,'' in \emph{Proceedings of the Genetic and Evolutionary
  Computation Conference Companion}, 2021, pp. 277--278.

\bibitem{ng2014one}
S.~S. Ng, P.~W. Tse, and K.~L. Tsui, ``A one-versus-all class binarization
  strategy for bearing diagnostics of concurrent defects,'' \emph{Sensors},
  vol.~14, no.~1, pp. 1295--1321, 2014.

\bibitem{adnan2015one}
M.~N. Adnan and M.~Z. Islam, ``One-vs-all binarization technique in the context
  of random forest,'' in \emph{Proceedings of the european symposium on
  artificial neural networks, computational intelligence and machine learning},
  2015, pp. 385--390.

\bibitem{allwein2000reducing}
E.~L. Allwein, R.~E. Schapire, and Y.~Singer, ``Reducing multiclass to binary:
  A unifying approach for margin classifiers,'' \emph{Journal of machine
  learning research}, vol.~1, no. Dec, pp. 113--141, 2000.

\bibitem{liu1999simultaneous}
Y.~Liu and X.~Yao, ``Simultaneous training of negatively correlated neural
  networks in an ensemble,'' \emph{IEEE Transactions on Systems, Man, and
  Cybernetics, Part B (Cybernetics)}, vol.~29, no.~6, pp. 716--725, 1999.

\bibitem{abbass2003pareto}
H.~A. Abbass, ``Pareto neuro-evolution: Constructing ensemble of neural
  networks using multi-objective optimization,'' in \emph{The 2003 Congress on
  Evolutionary Computation, 2003. CEC'03.}, vol.~3.\hskip 1em plus 0.5em minus
  0.4em\relax IEEE, 2003, pp. 2074--2080.

\bibitem{garcia2005cooperative}
N.~Garc{\'i}a-Pedrajas, C.~Herv{\'a}s-Mart{\'i}nez, and D.~Ortiz-Boyer,
  ``Cooperative coevolution of artificial neural network ensembles for pattern
  classification,'' \emph{IEEE transactions on evolutionary computation},
  vol.~9, no.~3, pp. 271--302, 2005.

\bibitem{stanley2002evolving}
K.~O. Stanley and R.~Miikkulainen, ``Evolving neural networks through
  augmenting topologies,'' \emph{Evolutionary computation}, vol.~10, no.~2, pp.
  99--127, 2002.

\bibitem{hsu2002comparison}
C.-W. Hsu and C.-J. Lin, ``A comparison of methods for multiclass support
  vector machines,'' \emph{IEEE transactions on Neural Networks}, vol.~13,
  no.~2, pp. 415--425, 2002.

\bibitem{dietterich1994solving}
T.~G. Dietterich and G.~Bakiri, ``Solving multiclass learning problems via
  error-correcting output codes,'' \emph{Journal of artificial intelligence
  research}, vol.~2, pp. 263--286, 1994.

\bibitem{rokach2010pattern}
L.~Rokach, \emph{Pattern classification using ensemble methods}.\hskip 1em plus
  0.5em minus 0.4em\relax World Scientific, 2010, vol.~75.

\bibitem{bautista2012minimal}
M.~{\'A}. Bautista, S.~Escalera, X.~Bar{\'o}, P.~Radeva, J.~Vitri{\'a}, and
  O.~Pujol, ``Minimal design of error-correcting output codes,'' \emph{Pattern
  Recognition Letters}, vol.~33, no.~6, pp. 693--702, 2012.

\bibitem{scikit-learn}
F.~Pedregosa, G.~Varoquaux, A.~Gramfort, V.~Michel, B.~Thirion, O.~Grisel,
  M.~Blondel, P.~Prettenhofer, R.~Weiss, V.~Dubourg, J.~Vanderplas, A.~Passos,
  D.~Cournapeau, M.~Brucher, M.~Perrot, and E.~Duchesnay, ``Scikit-learn:
  Machine learning in {P}ython,'' \emph{Journal of Machine Learning Research},
  vol.~12, pp. 2825--2830, 2011.

\bibitem{asuncion2007uci}
A.~Asuncion and D.~Newman, ``Uci machine learning repository,'' 2007.

\bibitem{berger1999error}
A.~Berger, ``Error-correcting output coding for text classification,'' in
  \emph{IJCAI-99: Workshop on machine learning for information
  filtering}.\hskip 1em plus 0.5em minus 0.4em\relax Citeseer, 1999.

\bibitem{verma2019error}
G.~Verma and A.~Swami, ``Error correcting output codes improve probability
  estimation and adversarial robustness of deep neural networks,'' in
  \emph{Advances in Neural Information Processing Systems}, 2019, pp.
  8646--8656.

\bibitem{garcia2011empirical}
N.~Garc{\'i}a-Pedrajas and D.~Ortiz-Boyer, ``An empirical study of binary
  classifier fusion methods for multiclass classification,'' \emph{Information
  Fusion}, vol.~12, no.~2, pp. 111--130, 2011.

\bibitem{kong1995error}
E.~B. Kong and T.~G. Dietterich, ``Why error-correcting output coding works,''
  1995.

\bibitem{grandini2020metrics}
M.~Grandini, E.~Bagli, and G.~Visani, ``Metrics for multi-class classification:
  an overview,'' \emph{arXiv preprint arXiv:2008.05756}, 2020.

\end{thebibliography}

\end{document}